%% file: neurips_2022.tex
\documentclass{article}

\usepackage[final]{neurips_2022}
\renewcommand{\P}{\mathbb{P}}
\newcommand{\N}{\mathbb{N}}
\newcommand{\E}{\mathbb{E}}
\newcommand{\R}{\mathbb{R}}
\newcommand{\Z}{\mathbb{Z}}

\newcommand{\calF}{\mathcal{F}}
\newcommand{\calX}{\mathcal{X}}
\newcommand{\calY}{\mathcal{Y}}

\newcommand{\calG}{\mathcal{G}}
\newcommand{\calL}{\mathcal{L}}

\newcommand{\calT}{\mathcal{T}}

\newcommand{\calN}{\mathcal{N}}

\newcommand{\calD}{\mathcal{D}}
\newcommand{\Epsilon}{\mathcal{E}}
\newcommand{\Lap}{\mathrm{Lap}}
\newcommand{\BM}{\mathrm{BM}}
\newcommand{\LNR}{\mathrm{LNR}}
\newcommand{\AT}{\mathrm{AboveThreshold}}
\newcommand{\RAT}{\mathrm{ReducedAboveThreshold}}
\newcommand{\Aug}{\mathrm{Aug}}
\newcommand{\Alg}{\mathrm{Alg}}
\newcommand{\Cov}{\mathrm{Cov}}

\usepackage{color-edits}
\addauthor{sw}{blue}
\addauthor{jw}{red}
\addauthor{rr}{red}

\usepackage[utf8]{inputenc} 
\usepackage[T1]{fontenc}    
\usepackage{hyperref}       
\usepackage{url}            
\usepackage{booktabs}       
\usepackage{amsfonts}       
\usepackage{nicefrac}       
\usepackage{microtype}      
\usepackage{xcolor}         

\usepackage{amsmath}
\usepackage{amsthm}
\usepackage{algorithm}
\usepackage[noend]{algpseudocode}
\usepackage{subfig}
\usepackage{graphicx}
\usepackage{xcolor}
\usepackage{bbm}
\usepackage{kpfonts}

\usepackage{comment}

\newtheorem{theorem}{Theorem}[section]
\newtheorem{lemma}[theorem]{Lemma}

\newtheorem{prop}[theorem]{Proposition}
\newtheorem{corollary}[theorem]{Corollary}

\newtheorem{definition}[theorem]{Definition}

\title{Brownian Noise Reduction:\\ 
Maximizing Privacy Subject to Accuracy Constraints
}

\author{%
  Justin Whitehouse\\
  Carnegie Mellon University\\
  \texttt{jwhiteho@andrew.cmu.edu}\\
    \And
  Zhiwei Steven Wu\\
  Carnegie Mellon University\\
  \texttt{zstevenwu@cmu.edu}\\
  \And
  Aaditya Ramdas\\
  Carnegie Mellon University \\
  \texttt{aramdas@cmu.edu} \\
  \And
  Ryan Rogers\\
  LinkedIn\\
  \texttt{rrogers@linkedin.com}\\
}

\hypersetup{draft}
\begin{document}

\maketitle

\input{abs}
\input{int}

\input{back}
\input{bm2}
\input{lnr2}
\input{rat}
\input{exp}

\input{conc}
\input{ack2}

\bibliography{bib.bib}{}
\bibliographystyle{plainnat}

\newpage

\newpage

\appendix

\input{appendices/martingale}
\input{appendices/bm_proofs2}
\input{appendices/rat_proof}

\input{appendices/additional_NR}

\input{appendices/bdry}

\end{document}

%% file: abs.tex
\begin{abstract}
There is a disconnect between how researchers and practitioners handle privacy-utility tradeoffs. Researchers primarily operate from a \textit{privacy first} perspective, setting strict privacy requirements and minimizing risk subject to these constraints. Practitioners often desire an \textit{accuracy first} perspective, possibly satisfied with the greatest privacy they can get subject to obtaining sufficiently small error. \citet{ligett2017accuracy} have introduced a ``noise reduction'' algorithm to address the latter perspective. The authors show that by adding correlated Laplace noise and progressively reducing it on demand, it is possible to produce a sequence of increasingly accurate estimates of a private parameter while only paying a privacy cost for the least noisy iterate released. In this work, we generalize noise reduction to the setting of Gaussian noise, introducing the \textit{Brownian mechanism}. The Brownian mechanism works by first adding Gaussian noise of high variance corresponding to the final point of a simulated Brownian motion. Then, at the practitioner's discretion, noise is gradually decreased by tracing back along the Brownian path to an earlier time.
Our mechanism is more naturally applicable to the common setting of bounded $\ell_2$-sensitivity, empirically outperforms existing work on common statistical tasks, and provides customizable control of privacy loss over the entire interaction with the practitioner. We complement our Brownian mechanism with $\RAT$, a generalization of the classical $\AT$ algorithm that provides adaptive privacy guarantees. Overall, our results demonstrate that one can meet utility constraints while still maintaining strong levels of privacy.
\end{abstract}

%% file: int.tex
\section{Introduction}
\label{sec:int}
Over the past decade, differential privacy has seen industry-wide adoption as a means of protecting sensitive information \citep{erlingsson2014rappor, greenberg2016apple}. By injecting appropriate amounts of noise, differentially private algorithms allow the computation of population-level quantities of interest while guaranteeing individual-level privacy.
Of the private mechanisms used in industry, those relating to private empirical risk minimization (ERM) are perhaps the most impactful, in part due to their application in machine learning tasks \citep{abadi2016deep, song2013stochastic}. Researchers have developed many private ERM mechanisms, ranging from least squares minimzation \citep{sheffet2017differentially, chaudhuri2011differentially} to subsampled gradient descent~\citep{abadi2016deep, balle2018improving, wang2019subsampled}.
Despite this vast literature, most existing results take the same broad approach: they aim to minimize error (statistical risk) subject to strict privacy guarantees. While this strict adherence to privacy constraints may be necessary in some applications, it often provides weak utility guarantees \citep{fienberg2010differential} and can make some learning tasks impossible \citep{dwork2009complexity}. 
Industry applications of differential privacy may desire an \textit{accuracy first} perspective, setting desired risk requirements for models used in production. Privacy may still be a desirable aspect of computation, but it is by no means the only goal; minimizing risk may take center stage. 

The main existing approach to this accuracy-oriented perspective on privacy was given by \citet{ligett2017accuracy}. These authors introduce a \textit{noise reduction mechanism} for gradually releasing a private, high-dimensional parameter. By leveraging a Laplace-based Markov process \citep{koufogiannis2015gradual}, they construct a mechanism for which the privacy loss of releasing arbitrarily many estimates of a parameter only depends on the privacy loss of the least noisy parameter viewed. This is in contrast to results about the composition of private algorithms, in which privacy degrades according to the total number of parameters witnessed \citep{dwork2010boosting, kairouz2015composition, murtagh2016complexity}. The authors also demonstrate how to privately query the utility of observed parameters on private data by coupling their Laplace-based mechanism with $\AT$, a classical differentially private algorithm~\citep{dwork2014algorithmic, lyu2016understanding}.

While the above mechanism provides significant privacy loss savings over a baseline method that doubles the privacy loss each round, Laplace noise is unfit for many settings in which $\ell_2$-sensitivity is used for calibrating noise. Since converting from $\ell_2$-sensitivity to $\ell_1$-sensitivity\footnote{The $\ell_p$  sensitivity of $f$ is defined as $\sup_{x \sim x'}||f(x) - f(x')||_p$ for $p \geq 1$.} incurs a dimension-dependent cost, it is important to develop a noise reduction technique with Gaussian noise.

\textbf{Contributions and paper outline.}
We introduce the \textit{Brownian mechanism}, a novel approach for privately releasing a parameter vector subject to accuracy constraints. The Brownian mechanism adds correlated Gaussian noise to a risk-minimizing parameter through a Brownian motion. Noise is then iteratively stripped by moving adaptively backwards along the random walk until a suitable stopping condition is met,  such as meeting a target accuracy on a public dataset. In Section~\ref{sec:bm}, we define the Brownian mechanism and characterize its privacy loss. Using machinery from martingale theory, we construct \textit{privacy boundaries} for the Brownian mechanism --- upper bounds on privacy loss that hold simultaneously with high probability. In particular, the failure probability of these bounds does not depend on the number of outcomes observed, overcoming a seeming need for a union bound faced by~\citet{ligett2017accuracy}. These privacy boundaries yield provable, high-probability bounds on privacy loss under data-dependent stopping conditions.

If private data is used to evaluate risk, then the data-dependent stopping conditions can themselves leak information. To counter this, we introduce $\RAT$ in Section~\ref{sec:rat}, a generalization of the classical $\AT$ algorithm for privately querying accuracy on sensitive data. We show how to couple $\RAT$ and the Brownian mechanism so that a data analyst only ever incurs \textit{twice} the privacy loss they would incur if they had queried accuracy on a public dataset. This is in contrast to the results in \citet{ligett2017accuracy}, which note that the privacy loss of $\AT$ often dominates the privacy loss incurred from using noise reduction.

\begin{figure}[h!]
    \centering
    \subfloat{
        \includegraphics[width=0.7\textwidth]{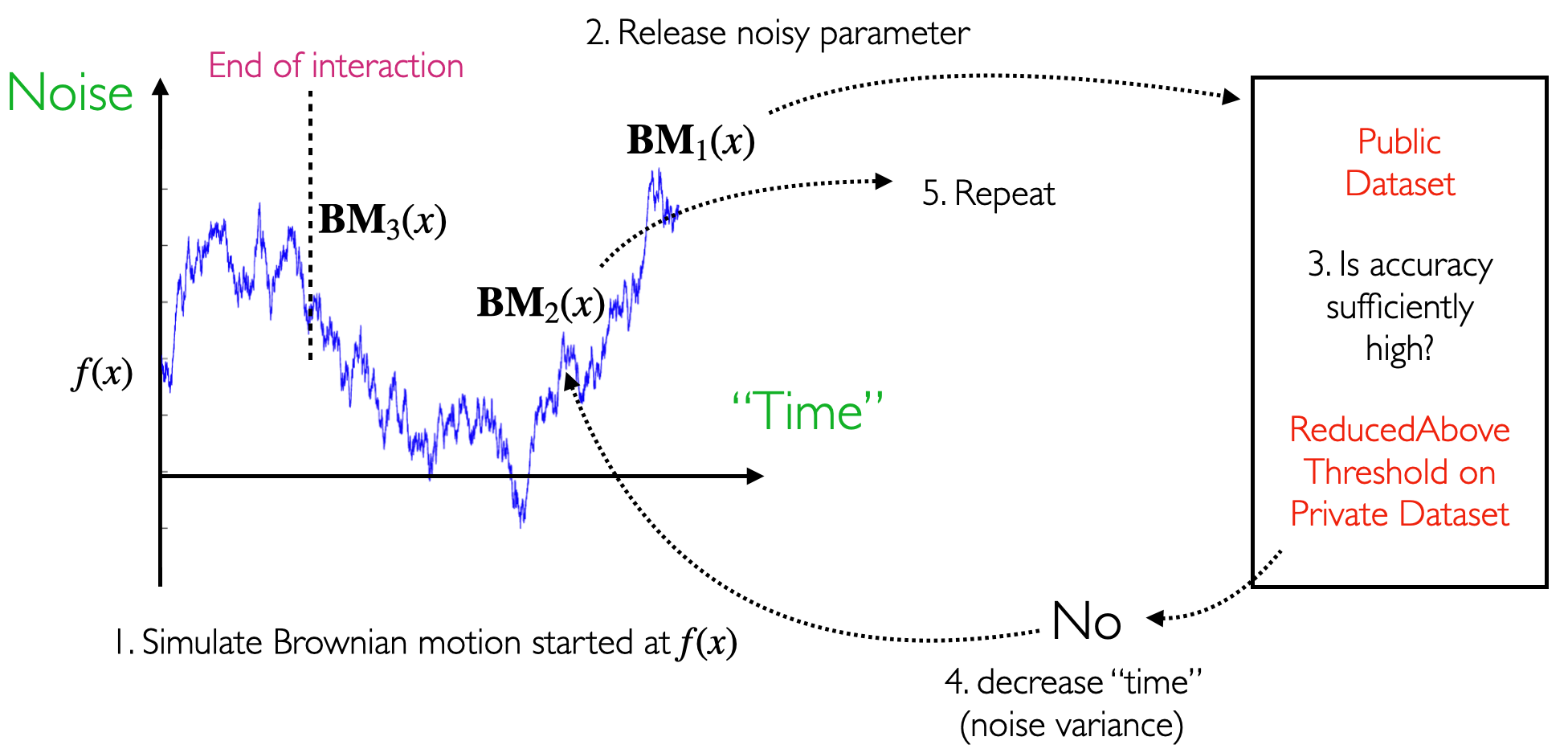}
    }
    \caption{An example of running the Brownian mechanism to gradually release a statistic $f(x)$. First, a very noisy version of the hidden parameter $\BM_{1}(x)$ is viewed. Then, loss is measured, either on a public dataset, or on a private dataset using a method such as $\RAT$. If a target loss is met, the process stops. Otherwise, noise is removed and the process repeats. }
    \label{fig:BM-cartoon}
\end{figure}

We empirically evaluate the Brownian mechanism and $\RAT$ in Section~\ref{sec:exp}, finding that the Brownian mechanism can offer privacy loss savings over the Laplace noise reduction method introduced by \citet{ligett2017accuracy}. In our view, these results demonstrate that the Brownian mechanism is a practical, intuitive mechanism for meeting accuracy requirements in private ERM.

Lastly,  we derive other new mechanisms for noise reduction, of independent interest. We generalize the Laplace process of \citet{koufogiannis2015gradual} to continuous time in Section~\ref{sec:lnr}, thus making the Laplace noise reduction mechanism of \citet{ligett2017accuracy} more flexible and adaptive to data-dependent privacy levels. We also briefly mention a noise reduction mechanism for Skellam noise in Section~\ref{sec:lnr}, a discrete distribution used in count queries~\citep{agarwal2021skellam}.

%% file: back.tex
\section{Preliminaries}
\label{sec:back}
\textbf{Differential privacy, privacy loss, and ex-post privacy.}
An algorithm  $A : \calX \rightarrow \calY$ is $(\epsilon, \delta)$-differentially private if, for any measurable set $E \subset \calY$ and any neighboring inputs $x \sim x'$, 
\begin{equation}
\label{eq:dp}
\P(A(x) \in E) \leq e^\epsilon \P(A(x') \in E) + \delta.
\end{equation}

In the above \citep{dwork2006priv}, $\sim$ denotes some arbitrary neighboring relation. Typically $x \sim x'$ indicates $x$ and $x'$ differ in one entry, but any other relation suffices. While differential privacy has proven itself a mainstay of private computation, condition~\eqref{eq:dp} is too rigid to allow data analysts to achieve a minimum desired accuracy. In other words, it embraces a \textit{privacy first} perspective, fixing a strict condition in terms of parameters $\epsilon$ and $\delta$ that must be met. We are interested in the \textit{accuracy first} perspective, setting a target accuracy and correspondingly optimizing privacy parameters.


The above definition of differential privacy is qualitatively focused on bounding the information-theoretic quantity of \textit{privacy loss}~\citep{dwork2006priv, dwork2010boosting, dwork2014algorithmic}.

\begin{definition}[\textbf{Privacy Loss}]
\label{def:ploss}
Let $A : \calX \rightarrow \calY$ be an algorithm, and fix neighbors $x \sim x'$ in $\calX$. Let $p^x$ and $p^{x'}$ be the respective densities of $A(x)$ and $A(x')$ on the space $\calY$ with respect to some reference measure\footnote{For instance, if $\mu_x$ and $\mu_{x'}$ are the laws of $A(x)$ and $A(x')$ respectively, the reference measure can be taken to be $\mu_x + \mu_{x'}$.}. Then, the privacy loss between $A(x)$ and $A(x')$ is the random variable
\begin{equation*}
\calL(x, x') := \log\left(\frac{p^x(A(x))}{p^{x'}(A(x))}\right).
\end{equation*}
\end{definition}

We think of $A(x)$ as the true outcome, and $\calL(x, x')$ measures how much more likely this outcome is under the true input $x$ versus an alternative $x'$. Privacy loss provides a \textit{probabilistic} definition of privacy. Namely, $A$ is \textit{$(\epsilon, \delta)$-probabilistically differentially private} if, for all neighbors $x \sim x'$,
\begin{equation}
\label{eq:pdp}
\P\left(\calL(x, x') > \epsilon\right) \leq \delta.
\end{equation}

While probabilistic differential privacy is not equivalent to differential privacy~\citep{kasiviswanathan2014semantics}, $(\epsilon, \delta)$-probabilistically differential privacy implies $(\epsilon, \delta)$-differential privacy. Probabilistic differential privacy emerged as a means for studying privacy composition, and has been leveraged in proving many results~\citep{kairouz2015composition, murtagh2016complexity, rogers2016odometer, whitehouse2022fully}. A natural extension of privacy to the accuracy-oriented regime is \textit{ex-post} privacy, which allows the bound in condition~\eqref{eq:pdp} to depend the observed algorithm output.

\begin{definition}[\citet{ligett2017accuracy}]
\label{def:expost}
Let $A : \calX \rightarrow \calY$ be an algorithm and $\Epsilon : \calY \rightarrow \R_{\geq 0}$ a function. We say $A$ is $(\Epsilon, \delta)$-ex-post private if, for any neighboring inputs $x \sim x'$, we have 
$$
\P\left(\calL(x, x') > \Epsilon(A(x))\right) \leq \delta.
$$
\end{definition}

While any algorithm is trivially ex-post private with $\Epsilon(A(x)) := \infty$, the goal is to make $\Epsilon(A(x))$ as small as possible. We describe theoretical tools for obtaining ex-post privacy guarantees in Section~\ref{sec:bm}, and empirically compute the ex-post privacy distributions of various mechanisms in Section~\ref{sec:exp}.


\textbf{Background on Noise Reduction.}
Heuristically, a noise reduction mechanism allows a data analyst to view multiple, increasingly accurate estimates of a risk minimizing parameter while only paying an ex-post privacy cost for the \textit{least} noisy iterate observed. Pinning down a general definition of a noise reduction mechanism is difficult, as any definition would need to depend on how the released parameter estimates were produced. In this paper, we consider the relevant case of additive noise mechanisms. Below, we provide an explicit definition of noise reduction mechanisms for this setting.

In the following definition, we let $(A_t)_{t \geq 0}$ be some collection of potentially correlated noise variables. In particular, $A_t$ should be thought of as marginally having either a multivariate normal distribution $\calN(0, t I_d)$ or multivariate Laplace distribution $\Lap(t)$. The index $t$ can be viewed as either ``time'' or ``variance'', with larger values of $t$ indicating greater variance of noise added. Further, when we refer to a sequence of \textit{time functions} $(T_n)_{n \geq 1}$, we mean a sequence of functions $T_n : (\R^d)^{n - 1} \rightarrow \R_{> 0}$ such that, for all $n \geq 1$ and $\beta_{1:n} \in (\R^d)^n$,
\begin{equation}\label{eq:time-functions}
T_{n + 1}(\beta_{1:n}) \leq T_n(\beta_{1:n - 1}).
\end{equation}
Intuitively, the $n$th time function gives the adaptively chosen variance of noise that will be added to the $n$th parameter based on the first $n - 1$ observed parameters. 

Let $M : \calX \rightarrow \calY^\infty$ be an algorithm mapping databases for sequences of outputs. Let $M_n : \calX \rightarrow \calY$ give the $n$th element of the sequence and $M_{1:n} : \calX \rightarrow \calY^n$ the first $n$ elements. We assume $M_n(x) := f(x) + A_{T_n(x)}$, where $f : \calX \rightarrow \calY$ is some function that should be thought of as producing a true, risk-minimizing parameter, $(T_n)_{n \geq 1}$ is a sequence of time functions, and $T_n(x) := T_n(M_{1: n -1}(x))$.

\begin{definition}[\textbf{Noise Reduction Mechanism}]
\label{def:nr}
Let $(A_t)_{t \geq 0}$ and $M : \calX \rightarrow \calY^\infty$ be as above, $a \in \calY$ any constant, and suppose $A_t + a$ has marginal density $p_t^a$.  We say $M$ is a \textit{noise reduction} mechanism if, for any $n \geq 1$ and any neighboring datasets $x \sim x'$, we have
\[
\calL_{1:n}(x, x') = \frac{p^{f(x)}_{T_n(x)}(M_n(x))}{p^{f(x')}_{T_n(x)}(M_n(x'))},
\]

where $\calL_{1:n}(x, x')$ denotes the privacy loss between $M_{1:n}(x)$ and $M_{1:n}(x')$. \end{definition}

The only noise reduction mechanism in the literature uses a Markov process with Laplace marginals~\citep{koufogiannis2015gradual} to gradually release a sensitive parameter~\citep{ligett2017accuracy}. As originally presented, this \textit{Laplace Noise Reduction} mechanism is nonadaptive, requiring a data analyst to fix a finite sequence of privacy parameters $(\epsilon_n)_{n \in [K]}$ in advance. Instead of presenting this method as background, we describe it in Section~\ref{sec:lnr}, in which we construct an adaptive generalization of this mechanism. We then leverage this generalization as a subroutine in $\RAT$, a generalization of $\AT$ with adaptive privacy guarantees.

\textbf{Background on Brownian Motion.}
We now provide a brief background on Brownian motion, perhaps the best-known example of a continuous time stochastic process~\citep{le2016brownian}.
\begin{definition}
\label{def:brownian}
A continuous time real-valued process $(B_t)_{t \geq 0}$ is called a standard Brownian motion if (1) $B_0=0$, (2) $(B_t)_{t \geq 0}$ has continuous sample paths, (3) $(B_t)$ has independent increments, i.e. $B_{t + s} - B_s$ is independent of $B_s$ for all $s, t \geq 0$, and (4) $B_t \sim \calN(0, t)$ for all $t \geq 0$.
\end{definition}

We say a process $(B_t)_{t \geq 0}$ is a $d$-dimensional standard Brownian motion if each coordinate process is an independent standard Brownian motion.

We use many properties of Brownian motion to construct the Brownian mechanism and analyze its privacy loss in Section~\ref{sec:bm}. One important property of Brownian motion is that it is a continuous time martingale. This property allow us to use time-uniform supermartingale concentration to characterize and bound the privacy loss of the Brownian mechanism at data-dependent stopping times \citep{howard2020line, howard2021unif}. We do not go into detail about martingale concentration in this background section, but rather defer it to  Appendix~\ref{app:martingale}. 
Additionally, $(B_t)_{t \geq 0}$ is a Markov process. This tells us that if we inspect the Brownian motion at times $0 \leq t_1 < t_2 < \cdots < t_n$, then $B_{t_2}, \dots, B_{t_n}$ can be viewed as a randomized post-processing of $B_{t_1}$ that \textit{does not} depend on $B_s$ for any $s < t_1$. This property allows us to show that the privacy loss of the Brownian mechanism --- which adds noise to a parameter via a Brownian motion --- only depends on the least noisy parameter observed.

%% file: bm2.tex
\section{The Brownian Mechanism: a Gaussian Noise Reduction Mechanism}
\label{sec:bm}

The Brownian mechanism works by simulating a Brownian motion starting at some multivariate parameter; this parameter should be thought of as the risk-minimizing output if there were no privacy constraints. The data analyst first observes the random walk at some large time.
Then, if so desired, the analyst ``rewinds'' time to an earlier point on the Brownian path, reducing noise to obtain a more accurate estimate. Due to the Markovian nature of Brownian motion, the analyst will only pay a privacy cost proportional to variance of the random walk at the earliest inspected time. 
\begin{equation}\label{eq:time-functions}
T_{n + 1}(\beta_{1:n}) \leq T_n(\beta_{1:n - 1}).
\end{equation}

\begin{definition}
\label{def:bm}
Let $f : \calX \rightarrow \R^d$ be a function and $(T_n)_{n \geq 1}$ a sequence of time functions. Let $(B_t)_{t \geq 0}$ be a standard $d$-dimensional Brownian motion. 
The Brownian mechanism associated with $f$ and $(T_n)_{n \geq 1}$ is the algorithm $\BM : \calX \rightarrow (\R^d)^\infty$ given by
\[
\BM(x) := \left(f(x) + B_{T_n(x)}\right)_{n \geq 1},
\]
where we set $T_n(x) := T_n\left(f(x) + B_{T_1(x)}, \dots, f(x) + B_{T_{n - 1}(x)}\right)$ with $T_1(x)$ being constant.

\end{definition}

We have chosen $T_n(x)$ as indexing notation to denote dependence on $x$, even if this is only through observed parameters. In the context of ERM, one can think of $f$ as computing a risk minimizing parameter associated with a private dataset $x \in \calX$. The data analyst uses $T_n$ along with the previous iterate to determine how far to rewind time to obtain the $n$th iterate. 

The Brownian mechanism, as defined above, produces an infinite sequence of parameters. In practice, a data analyst will only view finitely many iterates, stopping when some utility condition has been met or a minimum privacy level is reached. We introduce \textit{stopping functions} to model how a data analyst adaptively interacts with noise reduction mechanisms.

\begin{definition}[\textbf{Stopping Function}]
\label{def:stop}
Let $M : \calX \rightarrow \calY^\infty$ be a an algorithm. For $x \in \calX$, let $(\calF_n(x))_{n \in \N}$ be the filtration given by $\calF_n(x) := \sigma(M_i(x) :i \leq n)$.\footnote{The notation $\sigma(X)$ denotes the $\sigma$-algebra generated by $X$. $N$ is said to be a stopping time with respect to $(X_n)$ if $\{N \leq n\} \in \sigma(X_m :m \leq n)$ for all $n \in \N$. This definition can be extended to allow for $N$ to depend on independent, external randomization, but we omit this for simplicity.} A function $N : \calY^\infty \rightarrow \N$ is called a stopping function if for any $x \in \calX$, $N(x) := N(M(x))$ is a stopping time with respect to $(\calF_n(x))_{n \geq 1}$.
\end{definition}
A stopping function $N$ is a rule used to decide when to stop viewing parameters that \textit{only} depends on the observed iterates of the noise reduction mechanism. $N$ could heuristically be ``stop at the first time a parameter achieves an accuracy of 95\% on a held-out dataset.'' If a data analyst uses a stopping function alongside $\BM$, per Definition~\ref{def:nr}, the privacy loss accrued upon stopping is $\calL^{\BM}_{N(x)}(x, x')$.  
Recall from Figure~\ref{fig:BM-cartoon} and equation~\eqref{eq:time-functions} that the later iterations of BM correspond to smaller noise variances, meaning that $T_n$ is a decreasing sequence in the number of iterations $n$. Further, the filtration $\calF$ defined above is quite different from the usual filtrations considered for Brownian motions. 
In some cases, an analyst may want the stopping function to depend on the underlying private dataset through more than just the released parameters, e.g. they may want their rule to be ``stop at the first time a parameter achieves an accuracy of 95\% on the private dataset.'' In this case, additional privacy may be lost due to observing $N(x)$. We detail how to handle this more subtle case in Section~\ref{sec:rat}.

Due to the Markovian nature of Brownian motion, we get the following lemma. We include a proof in Appendix~\ref{app:bm} for completeness.

\begin{lemma}
\label{lem:bm_nr}
Let $x \sim x'$ be neighbors and $(T_n)_{n \geq 1}$ a sequence of time functions. Then, for any $n \geq 1$, letting $\calL_{1:n}^\BM(x, x')$ denote the privacy loss between $\BM_{1:n}(x)$ and $\BM_{1:n}(x')$, we have
\[
\calL_{1:n}^{\BM}(x, x') = \log\left(\frac{p_{T_n(x)}^{f(x)}(\BM_n(x))}{p_{T_n(x)}^{f(x')}(\BM_n(x))}\right),
\]
where $p^{\mu}_t$ is the density of a $\calN(\mu, t I_d)$ random variable. Furthermore, the above equality holds if $n$ is replaced by an almost surely bounded stopping function $N(x)$.

\end{lemma}

Lemma~\ref{lem:bm_nr} just tells us that the Brownian mechanism is a noise reduction mechanism, i.e.\ that the privacy lost by viewing the first $n$ iterates is exactly the privacy lost by viewing the $n$th iterate in isolation. 
Thus, we can identify $\calL_{1:n}^{\BM}(x, x')$ with $\calL_n^{\BM}(x, x')$ going forward.


The following theorem characterizes the privacy loss of the Brownian mechanism.
\begin{theorem}
\label{thm:bm_ploss}
Let $\BM$ be the Brownian mechanism associated with $(T_n)_{n \geq 1}$, a function $f: \calX \to \R^d$, and stopping function $N$. For neighbors $x \sim x'$,  the privacy loss between $\BM_{1:N(x)}(x)$ and $\BM_{1:N(x')}(x')$ is given by
\[
\calL_{1:N(x)}^{\BM}(x, x') =  \frac{||f(x) - f(x')||_2^2}{2T_{N(x)}(x)} + \frac{||f(x) - f(x')||_2}{T_{N(x)}(x)} W_{T_{N(x)}(x)},
\]
where $(W_t)_{t \geq 0}$ is a standard, univariate Brownian motion. Suppose $f$ has $\ell_2$-sensitivity at most $\Delta_2$. Then, letting $a^+ := \max(0, a)$, we have
\[
\calL_{1:N(x)}^{\BM}(x, x')  \leq \frac{\Delta_2^2}{2T_{N(x)}(x)} + \frac{\Delta_2}{T_{N(x)}(x)}W_{T_{N(x)}(x)}^+.
\]

\end{theorem}

Theorem~\ref{thm:bm_ploss} also holds when a deterministic time $n$ is replaced by  $N(x)$, where $N$ is a stopping function. 
The above theorem can be viewed as a process-level equivalent of the well-known fact that the privacy loss of the Gaussian mechanism has an uncentered Gaussian distribution~\citep{balle2018improving}. We prove the Theorem~\ref{thm:bm_ploss} in Appendix~\ref{app:bm}. Given the clean characterization of privacy loss above, we now show how to construct high-probability, time-uniform privacy loss bounds. We define \textit{privacy boundaries}, which map the variance of $\BM$ to high-probability bounds on privacy loss.

\begin{definition}
\label{def:priv_bound}
A function $\psi : \R_{\geq 0} \rightarrow \R_{\geq 0}$ is a \textit{$\delta$-privacy boundary} for the Brownian mechanism associated with time functions $(T_n)_{n \geq 1}$ if for any neighboring datasets $x \sim x'$, we have

\[
\P\left(\exists n \geq 1 : \calL_{1:n}^{\BM}(x, x') \geq \psi(T_n(x))\right) \leq \delta
\]

\end{definition}

 Since the privacy loss of $\BM$ is a deterministic function of a Brownian motion, we can apply results from martingale theory to construct general families of privacy boundaries.

\begin{theorem}
\label{thm:priv_bound}
Assume the same setup as in Theorem~\ref{thm:bm_ploss}. Let $\delta > 0$ and $f$ be a function with $\ell_2$-sensitivity $\Delta_2$. The following classes of functions form $\delta$-privacy boundaries.
\begin{enumerate}
\item{\textbf{(Mixture boundary)}} For any $\rho > 0$, $\psi^{M}_\rho$ given by
\[
\psi^{M}_\rho(t) := \frac{\Delta^2_2}{2t} + \frac{\Delta_2}{t}\sqrt{2(t + \rho)\log\left(\frac{1}{\delta}\sqrt{\frac{t + \rho}{\rho}}\right)}.
\]
\item{\textbf{(Linear boundary)}} For any $a, b > 0$ such that $2ab = \log(1/\delta)$, $\psi^{L}_{a, b}$ given by
\[
\psi^{L}_{a, b}(t) := \frac{\Delta_2}{t}\left(\frac{\Delta_2}{2} + b\right) + \Delta_2 a.
\]
\end{enumerate}

\end{theorem}

We prove Theorem~\ref{thm:priv_bound} in Appendix~\ref{app:bm}. In the same appendix, we plot the boundaries in Figure~\ref{fig:bound}. 

Privacy boundaries serve a dual purpose for the Brownian mechanism. First, since time-uniform concentration bounds are valid at arbitrary data-dependent times, that need not be stopping times with respect to the standard forward Brownian Motion filtration \citep{howard2021unif}, privacy boundaries provide ex-post privacy guarantees.  Second, in many settings, it may be more natural for a data analyst to adaptively specify target privacy levels instead of noise levels. 
This is, for instance, the case in our experiments in Section~\ref{sec:exp}. 
By inverting privacy boundaries, data analysts can compute the proper amount of noise to remove at each step to meet target privacy levels.

We make the above precise in Corollary~\ref{cor:ex_post}. In what follows, when we refer to a sequence $(\Epsilon_n)_{n \geq 1}$ of \textit{privacy functions}, we mean a sequence of functions $\Epsilon_n : (\R^d)^{n - 1} \rightarrow \R_{\geq 0}$ such that, for all $n$ and $\beta_{1:n} \in (\R^d)^{n}$, $\Epsilon_{n + 1}(\beta_{1:n}) \geq \Epsilon_n(\beta_{1:n-1})$. 

\begin{corollary}
\label{cor:ex_post}
Let $N$ be a stopping function, as in Definition~\ref{def:stop}. If $\psi$ is a $\delta$-privacy boundary for $\BM$, we have
\[
\sup_{x \sim x'}\P\left(\calL^{\BM}_{N(x)}(x, x') \geq \psi\left(T_{N(x)}(x)\right)\right) \leq \delta,
\]
i.e.\ the algorithm $\BM_{1:N(\cdot)}(\cdot)$ is $\left(\psi(T_{N(\cdot)}(\cdot)), \delta\right)$-ex post private, where $(\cdot)$ denotes a positional argument for an input $x \in \calX$. 
Further, let $(\Epsilon_n)_{n \geq 1}$ be a sequence of privacy functions, and define 
\[
T_n(\beta_{1:n - 1}) := \inf\left\{t \geq 0 : \psi(t) \geq \Epsilon_n(\beta_{1:n - 1})\right\}.
\]
Then $\BM_{1:N(\cdot)}(\cdot)$ is $(\Epsilon_{N(\cdot)}(\cdot), \delta)$-ex post private, where $\Epsilon_n(x)$ is defined analogously to $T_n(x)$.
\end{corollary}

Again, $N$ should be thought of as a stopping rule based on parameter accuracy. $\Epsilon_n$ should be thought of as a rule for choosing the $n$th privacy parameter given $\BM_{1:n-1}(x)$. 

%% file: lnr2.tex
\section{An Adaptive, Continuous-Time Extension of Laplace Noise Reduction}
\label{sec:lnr}

Here, we generalize the original noise reduction mechanism of \citet{ligett2017accuracy}, which will be used as a subroutine in Algorithm~\ref{alg:rat} in the following section. We first describe the original Laplace-based Markov process of \citet{koufogiannis2015gradual}. Fix any positive integer $K$ and any finite, increasing sequence of times $(t_n)_{n \in [K]}$. Let $(\zeta_n)_{n = 0}^K$ be the $d$-dimensional process given by $\zeta_0 = 0$ and
\begin{equation}
\label{eq:orig_lap_proc}
\zeta_n = \begin{cases}
\zeta_{n - 1} &\text{ with probability } \left(\frac{t_{n - 1}}{t_n}\right)^2 \\
\zeta_{n - 1} + \Lap(t_n) &\text{ otherwise.}
\end{cases}
\end{equation}

\citet{koufogiannis2015gradual} show that $\zeta_n \sim \Lap(t_n)$ and that $(\zeta_n)_{n = 0}^K$ is Markovian. \citet{ligett2017accuracy} use the above process to construct a noise reduction mechanism. Namely, they define the the \textit{Laplace Noise Reduction} mechanism associated with $f : \calX \rightarrow \R^d$ and $(t_n)_{n \in [K]}$ to be the algorithm $\LNR : \calX \rightarrow (\R^d)^K$ given by $\LNR(x) := (f(x) + \zeta_{K}, \dots, f(x) + \zeta_1)$. If $t_n := \Delta_1/\epsilon_n$, then releasing $n$th component $\LNR_n(x)$ in isolation is equivalent to running the classical Laplace mechanism with privacy level $\epsilon_n$. 

We now extend the process $(\zeta_n)_{n \in [K]}$ to a continuous time process with the same finite-dimensional distributions. Let $\eta > 0$ be arbitrary, and let $(P_t)_{t \geq \eta}$ be an inhomogeneous Poisson process with intensity function $\lambda(t) := \frac{2}{t}$. For $n \geq 1$, let $\calT_n := \inf\{t \geq \eta : P_t \geq  n\}$ be the $n$th jump of $(P_t)_{t \geq \eta}$ and set $\calT_0 := \eta$. Noting that $P_t$ must be a nonnegative integer, define the process $(Z_t)_{t \geq \eta}$ by
\begin{equation}
\label{eq:lap_proc}
Z_t :=  \sum_{n = 0}^{P_t}\Lap(\calT_n).
\end{equation}
It is immediate that $(Z_t)_{t \geq \eta}$ is Markovian. We show in Appendix~\ref{app:nr} that $Z_t \sim \Lap(t)$. With $(Z_t)_{t \geq \eta}$, one can make $\LNR$ fully adaptive, meaning that the times $(t_n)_{n \in [K]}$ at which it is invoked need not be prespecified, and can depend on the underlying input database $x$ by using time functions. 
\begin{definition}
\label{def:lnr}
Let $f : \calX \rightarrow \R^d$ be a function and $(T_n)_{n \geq 1}$ a sequence of time functions. Let $(Z_t)_{t \geq \eta}$ be the process defined in Equation~\eqref{eq:lap_proc}. The \textit{Laplace noise reduction} mechanism associated with $f$ and $(T_n)_{n \geq 1}$ is the algorithm $\LNR : \calX \rightarrow (\R^d)^\infty$ given by
\[
\LNR(x) := (f(x) + Z_{T_n(x)})_{n \geq 1},
\]

where again $T_n(x) := T_n(f(x) + Z_{T_1(x)}, \dots, f(x) + Z_{T_{n - 1}(x)})$ and $T_1(x)$ is constant.

\end{definition}

If the analyst would prefer instead to specify privacy functions $(\Epsilon_n)_{n \geq 1}$, they can do so by leveraging the corresponding time functions $T_n(x) := \Delta_1/\Epsilon_n(x)$, where $\Epsilon_n(x)$ is defined analogously to $T_n(x)$. We leverage $\LNR$ in our experiments in Section~\ref{sec:exp} and the process $(Z_t)_{t \geq 0}$ as a subroutine in constructing $\RAT$. An analogous argument to the one used in proving Lemma~\ref{lem:bm_nr} can be used to show $\LNR$ enjoys the following ex-post privacy guarantee.

\begin{prop}
\label{prop:lnr}
Let $\LNR$ be associated with $(T_n)_{n \geq 1}$ and a function $f$ with $\ell_1$-sensitivity $\Delta_1$. If $N$ is stopping function, the algorithm $\LNR_{1:N(\cdot)}(\cdot)$ is $(\Delta_1/T_{N(\cdot)}(\cdot), 0)$-ex post private.
\end{prop}


\paragraph{Skellam Noise Reduction.} Last, we briefly discuss how to generate a noise reduction mechanism for Skellam noise~\citep{agarwal2021skellam}. Recall that a random variable $X$ has a Skellam distribution with parameters $\lambda_1$ and $\lambda_2$ if $X =_d Y_1 - Y_2$, where $Y_1 \sim \mathrm{Poisson}(\lambda_1)$ and $Y_2 \sim \mathrm{Poisson}(\lambda_2)$ are independent Laplace random variables. For succinctness, we write $X \sim \textrm{Skell}(\lambda_1, \lambda_2)$.

Let $(P_1(t))_{t \geq 0}$ and $(P_2(t))_{t \geq 0}$ be two independent, homogeneous Poisson process with rates $\lambda_1$ and $\lambda_2$ respectively. Observe that the continuous time process $(X_t)_{t \geq 0}$ given by $X_t := P_1(t) - P_2(t)$ is clearly Markovian, has independent increments, and has $X_t \sim \mathrm{Skell}(t\lambda_1, t\lambda_2)$. Thus, $(X_t)_{t \geq 0}$ can be used to define a Skellam noise reduction mechanism by releasing $(f(x) + X_{T_n(x)})_{n \geq 1}$ for some sequence of time functions $(T_n)_{n \geq 1}$.

%% file: rat.tex
\section{Privately Checking if Accuracy is Above a Threshold}
\label{sec:rat}
In Section~\ref{sec:bm} we presented the Brownian mechanism, characterized its privacy loss, and showed how to obtain ex-post privacy guarantees for arbitrary stopping functions. In particular, these stopping functions could be based on the accuracy of the observed iterates on public held-out data.

However, one may desire to privately check the accuracy of observed iterates on the dataset $x \in \calX$.  \cite{ligett2017accuracy} were able to accomplish this goal by coupling $\LNR$ with $\AT$, a classical algorithm for privately answering threshold queries \citep{dwork2014algorithmic}. In the context of ERM, $\AT$ iteratively checks if the empirical risk of each parameter is below a target threshold, stopping at the first such occurrence. The downside to $\AT$ is that it requires a prefixed privacy level. In empirical studies, \citet{ligett2017accuracy} found this fixed privacy cost dominated the ex-post privacy guarantees, showing little benefit to using noise reduction.

Below, we construct $\RAT$, a generalization of $\AT$ which provides ex-post privacy guarantees. We show how to couple $\BM$ with $\RAT$ to obtain tighter ex-post privacy guarantees than coupling with $\AT$ would permit. In particular, if $\BM$ is run using parameters $(\epsilon_n)_{n \geq 1}$ and $\RAT$ indicates the $N$th parameter obtains sufficiently high accuracy, the privacy loss of the net procedure will be at most $2\epsilon_N$ --- only twice the privacy loss that would be accrued by testing on public data.

\begin{algorithm}[!h]
\caption{ReducedAboveThreshold (via Laplace Noise Reduction)}\label{alg:rat}
\begin{algorithmic}
\Require Algorithm $\Alg : \calX \rightarrow \calY^\infty$, parameter $\epsilon_{\max} > 0$, threshold $\tau$, database $x \in \calX$ , utility $u: \calY \times \calX \rightarrow \R$ where $u(\beta,\cdot)$ is $\Delta$-sensitive $\forall \beta$, privacy functions $(\Epsilon_n)_{n \geq 1}$ with $\Epsilon_n \leq \epsilon_{\max}$ $\forall n$.
\For{$n \geq 1$}
    \State $\epsilon_n := \Epsilon_n(\Alg_{1:n - 1}(x))$, $T_n := 2\Delta/\epsilon_n$
    \State $\zeta_n := Z_{T_n}$, where $(Z_t)_{t \geq \eta}$ in Eq.~\eqref{eq:lap_proc} defines the $\LNR$ mechanism with $\eta := 2\Delta/\epsilon_{\max}$.
    \State $\xi_n \sim \Lap\left(\frac{4\Delta}{\epsilon_n}\right)$
    \If{$u(\Alg_n(x), x) + \xi_n \geq \tau + \zeta_n$}
        \State Print $1$ and HALT
    \Else \State Print $0$
    \EndIf
\EndFor
\end{algorithmic}
\end{algorithm}

$\tau$ should be seen as a target accuracy, $\Alg$ as a mechanism for releasing a parameter (e.g.\ $\BM$, $\LNR$), and $u$ as evaluating the accuracy of $\Alg_n(x)$ on $x$. $\epsilon_{\max}$ is an arbitrarily large constant, representing the minimum level of privacy required, used to prevent the user from examining $(Z_t)$ at arbitrarily small times. The above generalizes to sequences of thresholds $(\tau_n)_{n \geq 1}$ and sequences $(u_n)_{n \geq 1}$ of functions $u_n : \calY^{n} \times \calX \rightarrow \R$ that are $\Delta$-sensitive in their second argument, but the added generality yields only marginal benefits.
When $\Epsilon_n = \epsilon$ for all $n$, Algorithm~\ref{alg:rat} recovers $\AT$ as a special case. The intuition behind $\RAT$ is that by gradually removing Laplace noise from the threshold, a data analyst can ensure that privacy of the whole procedure only depends on the magnitude of Laplace noise added when the algorithm halts. The following characterizes the privacy loss of Algorithm~\ref{alg:rat}.

\begin{theorem}
\label{thm:rat}
For any $n \geq 1$ and neighboring datasets $x \sim x'$, let $\calL^{\Alg}_{1:n}(x, x')$ denote the privacy between $\Alg_{1:n}(x)$ and $\Alg_{1:n}(x')$. For any $x \in \calX$, define $N(x)$ to be the first round where $\RAT$ run on input $x \in \calX$ outputs 1, that is
\[
N(x) := \inf\{n \geq 1 : \RAT_n(x) = 1\}.
\]
Then, the privacy loss between $\RAT(x)$ and $\RAT(x')$, denoted $\calL^{\mathrm{RAT}}(x, x')$, is bounded by
\[
\calL^{\mathrm{RAT}}(x, x') \leq  \calL^{\Alg}_{1:N(x)}(x, x') + \Epsilon_{N(x)}(\Alg_{1:N(x)-1}(x)).
\]
\end{theorem}

We prove Theorem~\ref{thm:rat} in Appendix~\ref{app:rat}, where we also provide a utility guarantee for $\RAT$. This utility guarantee, much like the utility guarantee for $\AT$, is in practice weak as it derives from a union bound.
Using Theorem~\ref{thm:rat}, we can simply choose $\Alg = \BM$ as a means of adaptively generating parameters. The following corollary, which follows immediately from the above theorem, provides the ex-post privacy guarantees of combining $\RAT$ and $\BM$.
\begin{corollary}
\label{cor:aug_bm}
Let $\BM$ be the Brownian mechanism associated with a function $f$, decreasing time functions $(T_n)_{n \geq 1}$, and a a $\delta$-privacy boundary $\psi$. Let $\RAT$ be run with privacy functions $(\psi(T_n))_{n \geq 1}$, threshold $\tau$, and algorithm $\BM$. Then, $\RAT$ is $\left(2\psi(T_{N(\cdot)}(\cdot)), \delta\right)$-ex post private.
\end{corollary}

%% file: exp.tex
\section{Experiments}
\label{sec:exp}
\begin{figure}[h]
    \centering
    \subfloat[Regularized Logistic Regression]{
        \includegraphics[width=0.4\textwidth]{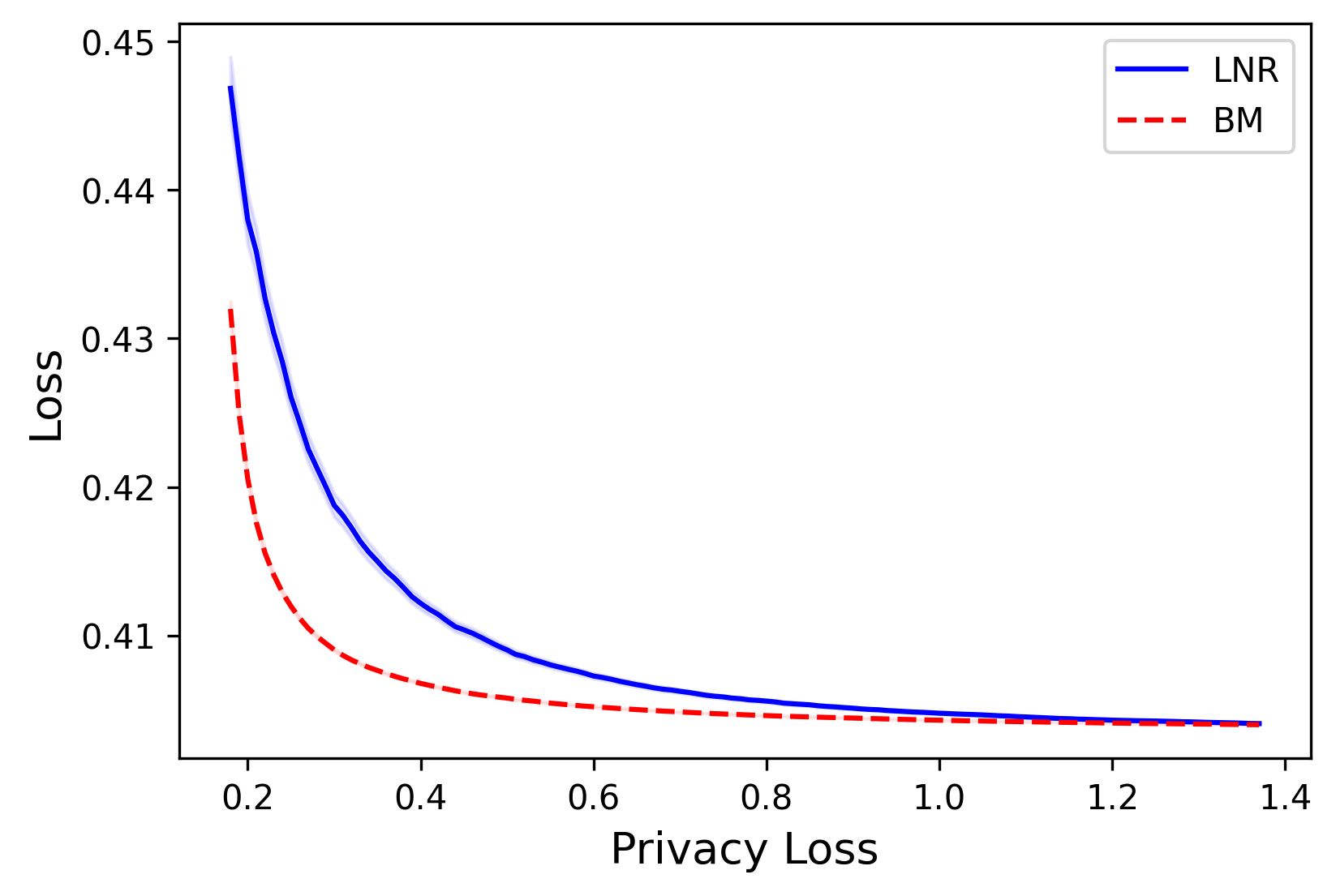}
        \label{fig:loss:lr}
    }
    \subfloat[Ridge Regression]{
        \includegraphics[width=0.42\textwidth]{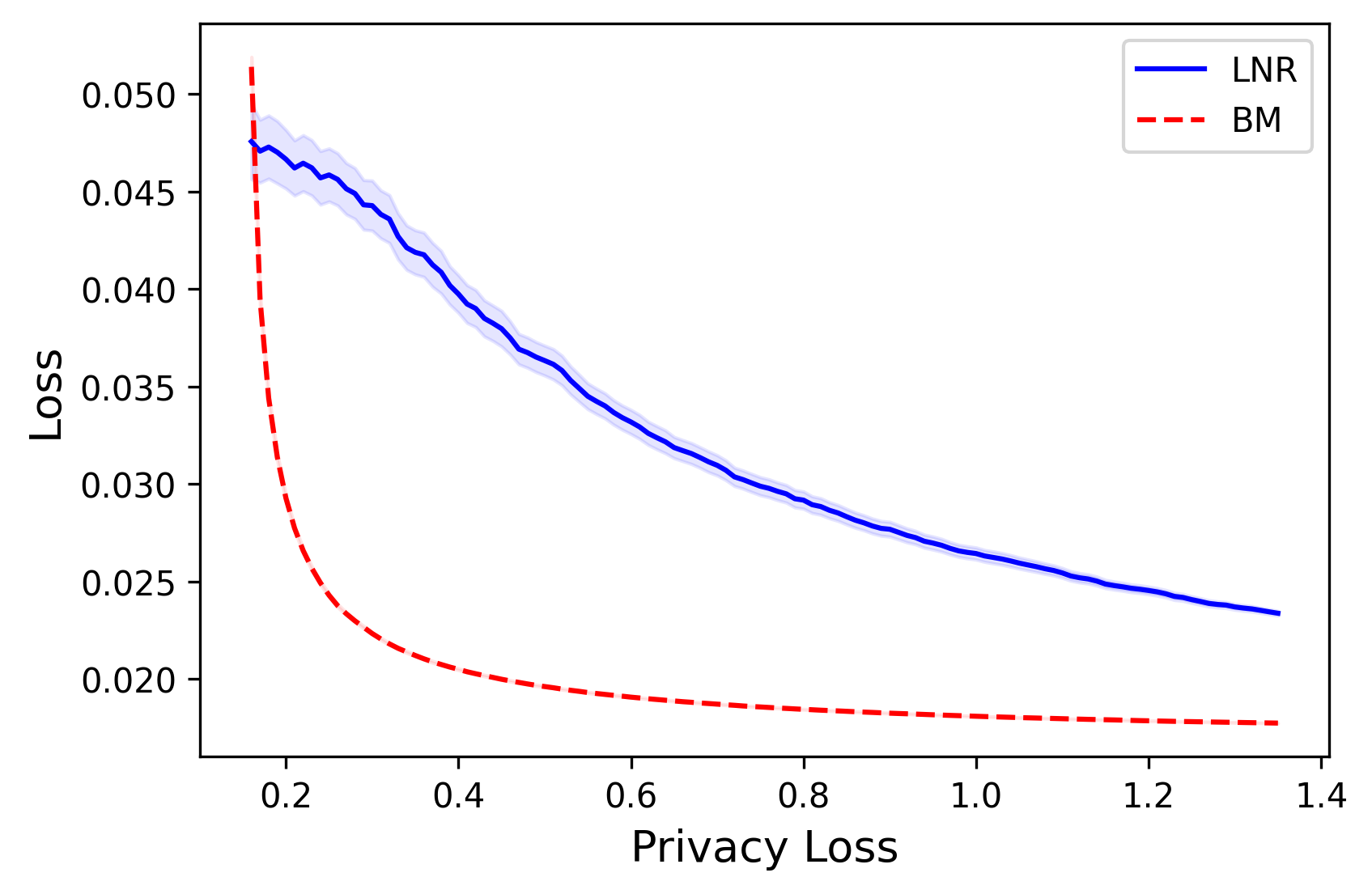}
        \label{fig:loss:rr}
    }
    
    \caption{
        Privacy loss plotted against loss (respectively regularized logistic and ridge loss)  for the statistical tasks of regularized logistic regression and ridge regression. 
    }
    \label{fig:loss}
\end{figure}

\begin{figure}[h]
    \centering
    \subfloat[Regularized Logistic Regression]{
        \includegraphics[width=0.4\textwidth]{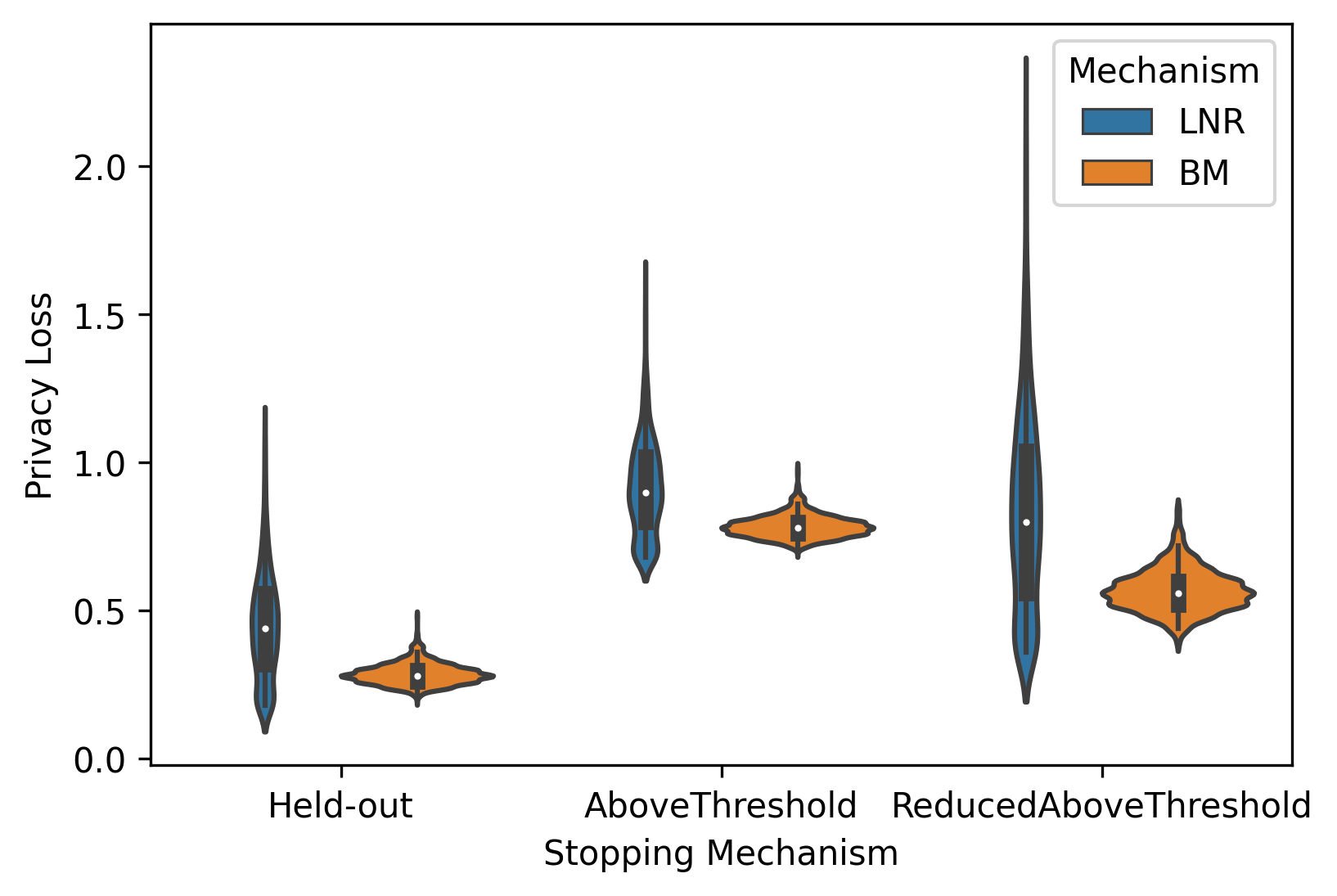}
        \label{fig:emp:lr}
    }
    \subfloat[Ridge Regression]{
        \includegraphics[width=0.4\textwidth]{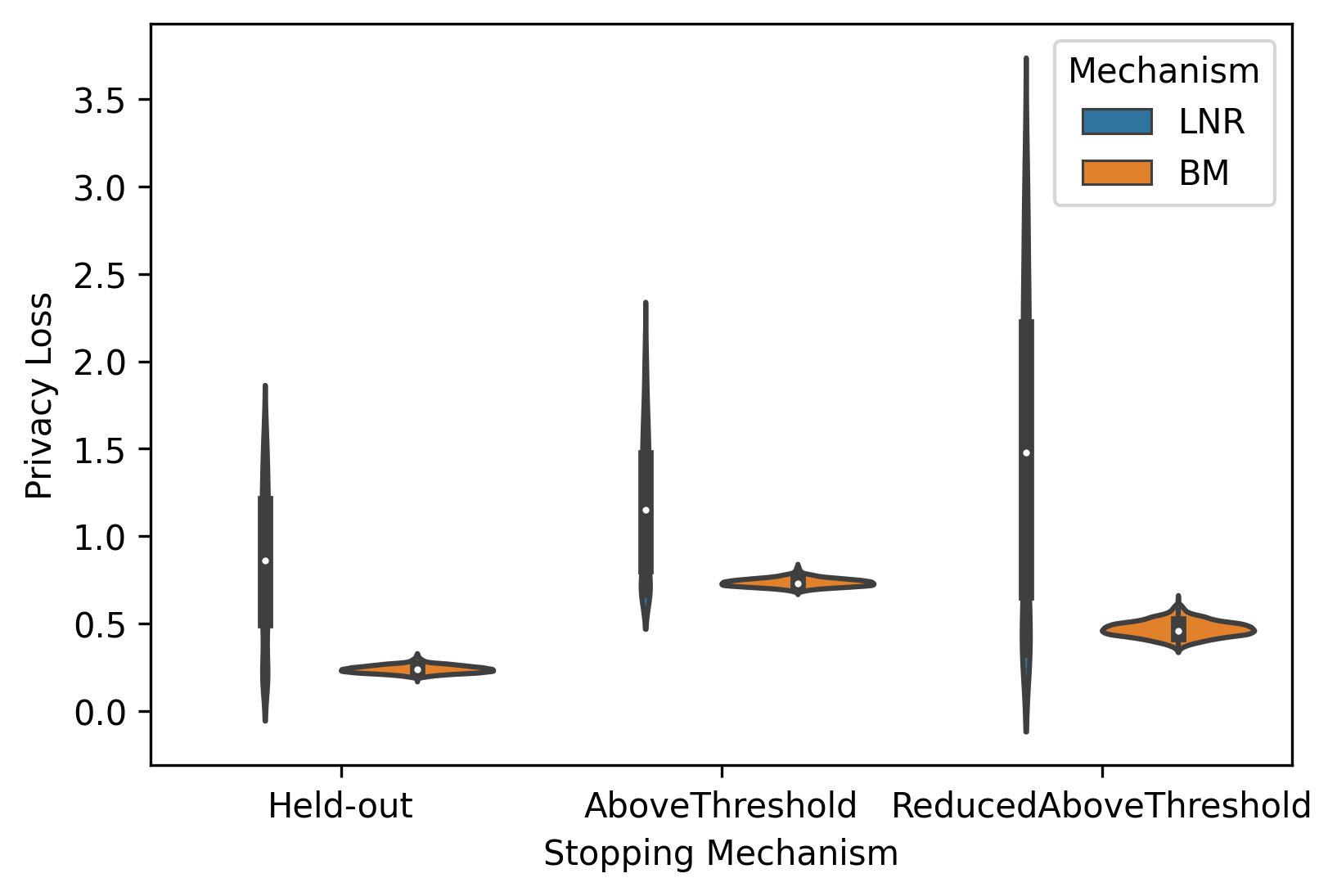}
        \label{fig:emp:rr}
    }

    \caption{Empirical privacy loss distributions for logistic regression and ridge regression with loss assessed either (left) on the training data treated as a public, held-out dataset, (middle) via $\AT$, or (right) via $\RAT$. 
    }
    \label{fig:emp}
\end{figure}

\textbf{Choice of tasks:} We compare the performance of $\BM$ and $\LNR$ on the tasks of regularized logistic regression via output perturbation~\citep{chaudhuri2011differentially} and ridge regression via covariance perturbation~\citep{smith2017interaction}.\footnote{The two tasks use the logistic loss  $\ell(y, z) := \log(1 + \exp(-yz))$ and the squared loss $\ell(y, z) := \frac{1}{2}(z - y)^2$. The regularized loss on a dataset $\calD := \{(x_i, y_i)\}_{i \in [n]}$ is 
\(
L(\beta, \calD) := \frac{1}{n}\sum_{i = 1}^n \ell(y_i, \beta^T x_i) + \frac{\lambda ||\beta||_2^2}{2}.
\)
}
For logistic regression, we leveraged the KDD-99 dataset~\citep{kdd99} with $d = 38$ features, predicting whether network events can be classified as ``normal" or ``malicious". For ridge regression, we used the Twitter dataset~\citep{kawala2013predictions} with $d = 77$ features to predict log-popularity of posts. In each case, we ran our experiments on $n = 10,000$ randomly sub-sampled data points. In order to guarantee bounded sensitivity, we normalized each data point to have unit $\ell_2$ norm. We note that this aspect differs from the experimentation conducted by \citet{ligett2017accuracy}, who normalized by the \textit{maximum} $\ell_2$ norm, a non-private operation.

\textbf{Experiments:} For each task, we conducted two experiments. We discuss the specific parameter settings for these experiments in Appendix~\ref{app:bdry}. In the first experiment, we plotted guaranteed (in the case of $\LNR$) or high-probability (in the case of $\BM$) privacy loss on the x-axis against average loss (either logistic or ridge) on the y-axis. We conduct such a comparison as probability 1 privacy loss bounds cannot be provided for the Gaussian mechanism. Likewise, adding a probability $\delta$ of minimally improves privacy loss for the Laplace mechanism. We computed the average loss curve for each mechanism over 1,000 trials, and have included point-wise valid 95\% confidence intervals.

In the second experiment, we plotted the empirical privacy loss distributions for $\BM$ and $\LNR$ under the stopping conditions of loss being at most 0.41 for logistic regression and 0.025 for ridge regression. For each mechanism, we evaluated this empirical distribution using three approaches for testing empirical loss: treating the training data as a held-out dataset, using $\AT$, and using our mechanism, $\RAT$.  In $\AT$, we set the privacy parameter to be fixed at $\epsilon = 0.5$. In $\RAT$, we took the sequence of privacy parameters to be the same as the sequence of privacy parameters used by $\BM$ and $\LNR$. We once again computed these empirical distributions over 1,000 runs of each mechanism.

\textbf{Findings:} The findings of the two experiments are summarized in Figure~\ref{fig:loss} and Figure~\ref{fig:emp}. For both tasks, $\BM$ obtains significant improvements in loss over $\LNR$ near the privacy loss level that was optimized for. For both tasks, the privacy loss distribution for $\BM$ has lower median privacy loss than that of $\LNR$. In addition, the privacy loss distribution for $\BM$ is more tightly concentrated around the median, indicating more consistent performance. The privacy loss distribution for $\LNR$ has a heavy tail, demonstrating that many runs do not attain the target loss until high privacy loss costs are incurred.   
Comparing $\RAT$ and $\AT$, we see that the privacy loss distribution for $\RAT$ has higher variance than that of $\AT$. However, $\RAT$ attains a significantly lower median level of privacy loss when coupled with $\BM$. This latter point reflects the observations of \citet{ligett2017accuracy}, who note that when $\AT$ is used to determine stopping conditions on private data, it contributes the bulk of the privacy loss to the empirical distributions. On the other hand, our figures demonstrate that $\RAT$ results in a more mild privacy loss at target stopping conditions.

%% file: conc.tex
\section{Conclusion}
\label{sec:conc}

In this paper, we constructed the Brownian mechanism ($\BM$), a novel approach to noise reduction that adds noise to a hidden parameter via a Brownian motion. We not only precisely characterized the privacy loss of the Brownian mechanism, but also bounded it through applying machinery from continuous time martingale theory. We then demonstrated how the utility of the iterates produced by $\BM$ can be assessed on private data via $\RAT$, a generalization of the classical $\AT$ algorithm. This was itself accomplished by a continuous-time generalization of the original Laplace noise reduction ($\LNR$) mechanism. Last, we empirically demonstrated that $\BM$ outperforms $\LNR$ on common statistical tasks, such as regularized logistic and ridge regression. 

We comment on several limitations and open problems related to our work. We considered noise reduction mechanisms in the setting of one-shot privacy, in which only a single mechanism is run on private data. Traditional composition results, such as those for fixed privacy parameters~\citep{dwork2010boosting, kairouz2015composition, murtagh2016complexity} or adaptively selected parameters~\citep{rogers2016odometer, feldman2021individual, whitehouse2022fully} are not directly applicable to algorithms satisfying ex-post privacy; additional machinery needs to be developed to handle composition in this case. A naive approach to composition is possible, which involves summing the ex-post privacy guarantees of composed algorithms and summing the corresponding $\delta$'s, but we expect this approach to be loose.
Finally, noise reduction is currently only applicable to output perturbation methods; it remains open to see how to combine noise reduction with other prominent methods for private computation, such as objective perturbation.


%% file: ack2.tex
\section{Acknowledgements}

We would like to thank Gennady Samorodnitsky for pointing out a mistake in an earlier version of Definition~\ref{def:nr}, which in turn led to a bug in the proof of Lemma~\ref{lem:bm_nr}.
AR acknowledges support from NSF DMS 1916320 and an ARL IoBT CRA grant. Research reported in this paper was sponsored in part by the DEVCOM Army Research Laboratory under Cooperative Agreement W911NF-17-2-0196 (ARL IoBT CRA). The views and conclusions contained in this document are those of the authors and should not be interpreted as representing the official policies, either expressed or implied, of the Army Research Laboratory or the U.S. Government. The U.S. Government is authorized to reproduce and distribute reprints for Government purposes notwithstanding any copyright notation herein.

ZSW and JW were supported in part by the NSF CNS2120667, a CyLab 2021 grant, a Google Faculty Research Award, and a Mozilla Research Grant.

JW acknowledges support from NSF GRFP grants DGE1745016 and DGE2140739.

%% file: appendices/martingale.tex
\section{Background on Martingale Concentration}
\label{app:martingale}

In this section, we provide a background on the basics of martingale concentration needed throughout this paper. Central to all results in this section is Ville's inequality \citep{ville1939etude}, which can be viewed as a time-uniform version of Markov's inequality for martingales.

\begin{lemma}[\textbf{Ville's Inequality} \citep{ville1939etude}]
\label{lem:ville}
Let $(X_t)_{t \geq 0}$ be a nonnegative supermartingale with respect to some filtration $(\calF_t)_{t \geq 0}$. Then, for any confidence parameter $\delta \in (0, 1)$, we have 
\[
\P\left(\exists t \geq 0: X_t \geq \frac{\E X_0}{\delta}\right) \leq \delta.
\]
\end{lemma}
While standard Brownian motion $(B_t)_{t \geq 0}$ is not a nonnegative supermartingale, geometric Brownian motion given by $Y_t^\lambda := \exp\left(\lambda B_t - \frac{\lambda^2}{2}t\right)$ is a nonnegative martingale for any $\lambda \in \R$, and hence Lemma~\ref{lem:ville} can be applied. In fact, the probability in the lemma above becomes exactly $\delta$ when it is applied to a nonnegative martingale with continuous paths like $Y_t^\lambda$. From Ville's inequality, the following \textit{line-crossing inequality} for Brownian motion can be obtained.

\begin{lemma}[Line-Crossing Inequality]
\label{lem:line_cross}
For $\delta \in (0, 1)$ and $a, b > 0$ satisfying $e^{-2ab} = \delta$, we have
\[
\P\left(\exists t \geq 0 : B_t \geq a t + b\right) = \delta.
\]

\end{lemma}

A proof of the above fact can be found in any standard book on continuous time martingale theory~\citep{le2016brownian, durrett2019probability}. The above also follows from a special case of the more general time-uniform Chernoff bound presented in \citet{howard2020line}. 

The above inequality can be seen as optimizing the tightness of the time-uniform boundary at one pre-selected point in time. However, due to the adaptive nature of the Brownian mechanism presented in Section~\ref{sec:bm}, it is sometimes desirable to construct a time-uniform boundary which sacrifices tightness at a fixed point in time to obtain greater tightness over all of time. 

The \textit{method of mixtures} provides one such approach for constructing tighter time-uniform boundaries~\citep{kaufmann2018mixture, howard2021unif}. We discuss this concept briefly in the context of Brownian motion. Observe that, since $(Y^\lambda_t)_{t \geq 0}$ is a nonnegative martingale, for any probability measure $\pi$ on $\R$, the process $(X^\pi_t)_{t \geq 0}$ given by 
\[
X^\pi_t := \int_\R Y^\lambda_t \pi(d\lambda)
\]
is also nonnegative martingale. By appropriately choosing the probability measure $\pi$ and applying Ville's inequality, one obtains the following concentration inequality~\citep{howard2021unif}.

\begin{lemma}[Mixture Inequality]
\label{lem:mixture}
Let $\rho > 0$ and $\delta \in (0, 1)$ be arbitrary. Then,
\[
\P\left(\exists t \geq 0 : B_t \geq \sqrt{2(t + \rho)\log\left(\frac{1}{\delta}\sqrt{\frac{t + \rho}{\rho}}\right)}\right) = \delta.
\]

\end{lemma}

We leverage Lemmas~\ref{lem:line_cross} and \ref{lem:mixture} to construct the privacy boundaries in Theorem~\ref{thm:priv_bound} in Appendix~\ref{app:bm}.

%% file: appendices/bm_proofs2.tex
\section{Proofs From Section~\ref{sec:bm}}
\label{app:bm}

\begin{figure}
\centering
    \subfloat[Variance of Noise vs. Privacy Loss]{
        \includegraphics[width=0.4\textwidth]{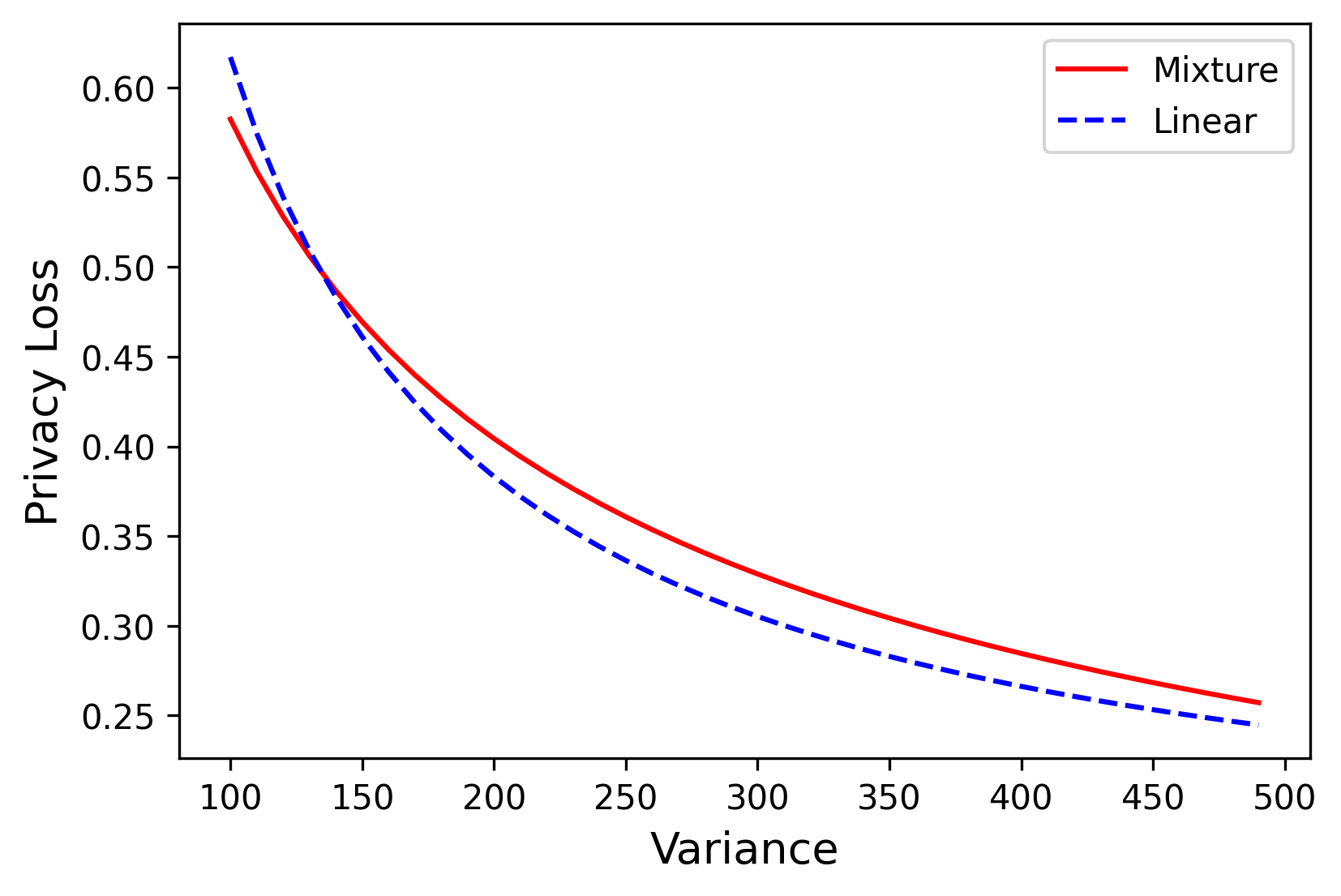}
        \label{fig:bound:var_pl}
    }
    \subfloat[Privacy Loss vs. Variance of Noise]{
        \includegraphics[width=0.4\textwidth]{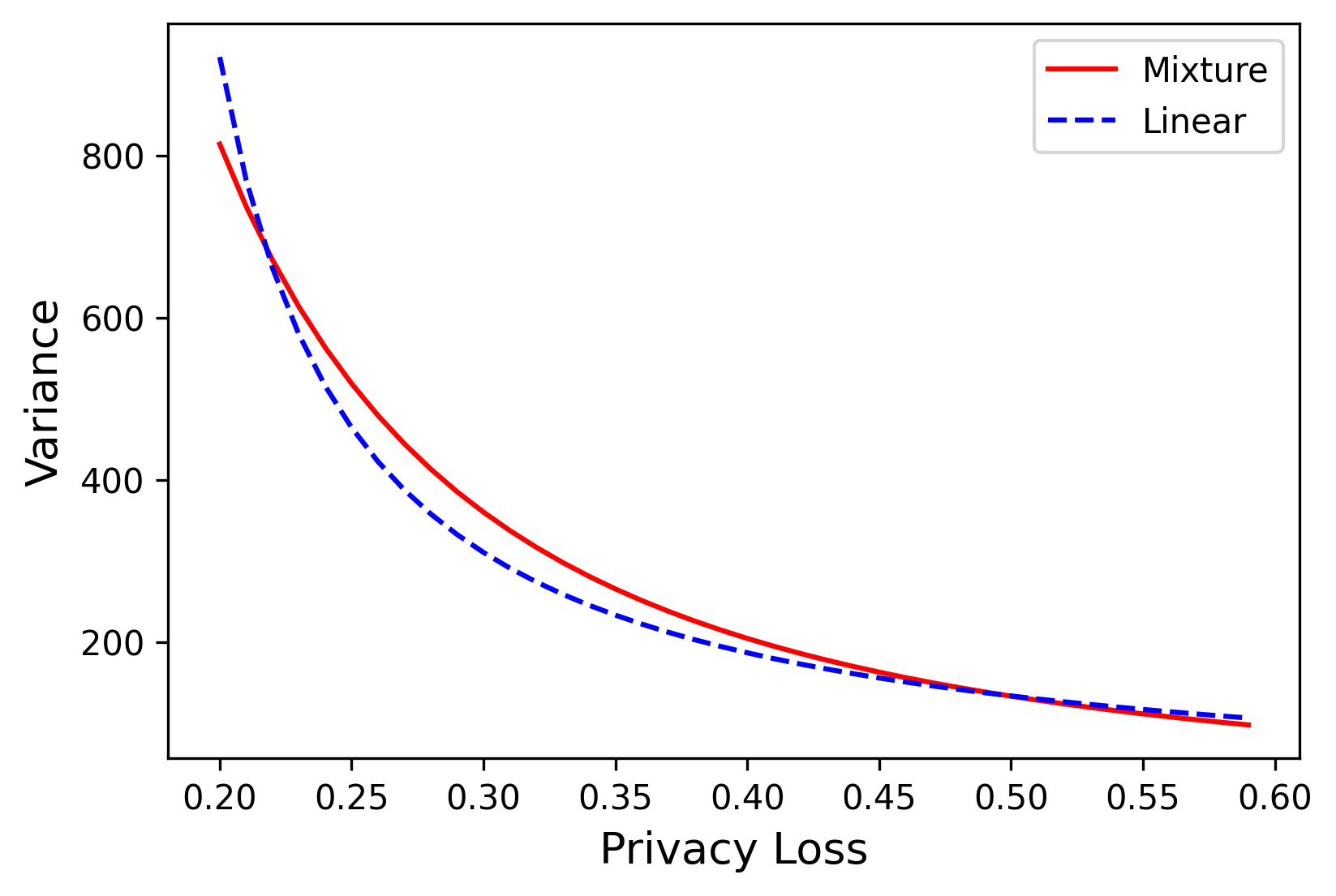}
        \label{fig:bound:pl_var}
    }
    
    \caption{
    A comparison of the linear and mixture boundaries, both optimized for tightness at $\epsilon = 0.3$ with $\delta = 10^{-6}$. The first plot directly plots the corresponding bounds as in Theorem~\ref{thm:priv_bound}. The second plot inverts the boundaries, showing the variance necessary to meet a target privacy level. 
    }
    \label{fig:bound}
\end{figure}

Here, we prove the results from Section~\ref{sec:bm}. We start by showing that $\BM$ is in fact a noise-reduction mechanism, which is claimed in Lemma~\ref{lem:bm_nr}. To prove the cited lemma, it suffices to show the following result.

\begin{prop}
\label{prop:target_ploss}
For $\nu \in \R^d$, let $(B_t^\nu)_{t \geq 0}$ be a standard $d$-dimensional Brownian motion starting at $\nu$. Let $(T_n)_{n \geq 1}$ be a sequence of decreasing time functions\footnote{As before, $T_1$ is implicitly a constant, independent of $(B_t^\nu)_{t \geq 0}$} $T_n : \R^{(n - 1)d} \rightarrow \R$, $N : \R^\infty \rightarrow \N$ a bounded stopping function, and define $T^\nu_n := T_n\left(B_{T_1^\nu}^\nu, \dots, B_{T_{n - 1}^\nu}^\nu\right)$ and $N^\nu := N\left((B^\nu_{T_n^\nu})_{n \geq 1}\right)$. Let $p^\nu_{1:N}$ denote the joint density of $\left(B^\nu_{T_1^\nu}, \dots, B^\nu_{T_{N^\nu}^\nu}\right)$. Then, with probability 1, we have
\[
\frac{p^\nu_{1:N}\left(B_{T_1^\nu}^\nu, \dots, B_{T_{N^\nu}^\nu}^\nu\right)}{p^\mu_{1:N}\left(B_{T_1^\nu}^\nu, \dots, B_{T_{N^\nu}^\nu}^\nu\right)} = \frac{\exp\left(-\frac{(B_{T_{N^\nu}^\nu}^\nu - \nu)^2}{2T_{N^\nu}}\right)}{\exp\left(-\frac{(B_{T_{N^\nu}^\nu}^\nu - \mu)^2}{2T_{N^\nu}}\right)},
\]
which is just the ratio between the density of a  $\calN\left(\nu, T_{N^\nu}^\nu\right)$ random variable and a $\calN\left(\mu, T_{N^\nu}^\nu\right)$ random variable evaluated at $B_{T_{N^\nu}^\nu}^\nu$. 
    
\end{prop}

A key part of proving the above proposition will be developing a strong Markov property for Brownian bridges. Recall that a \textit{Brownian bridge} is, in essence, a Brownian motion that has been ``pinned down'' at some initial and terminating value. More rigorously, for a random variable $A \in \R^d$ and a constant $b \in \R^d$, a Brownian bridge $(X_t)_{0 \leq t \leq T}$ with initial value $X_0 = A$ and terminating value $X_T = b$ is a process that can be written in the form $X_t = \frac{T - t}{T}A + B_t - \frac{t}{T}(B_T - b)$, where $(B_t)_{0 \leq t \leq T}$ is a standard $d$-dimensional Brownian motion that is independent of $A$. The following properties of Brownian bridges follow from the definition.

\begin{lemma}[Properties of Brownian Bridges]
\label{lem:bb_prop}
Let $(X_t)_{0 \leq t \leq T}$ be a $d$-dimensional Brownian bridge with $X_0 = A$, for $A$ being a random vector in $\R^d$, and $X_1 = b$, with $b \in \R^d$ fixed. Then, the following hold:
\begin{enumerate}
    \item If $A' \in \R^d$ is independent of $(X_t)_{t \geq 0}, A$ and $b' \in \R^d$ is constant, the process $(X_t')_{0 \leq t \leq T}$ given by
    \[
        X_t' := X_t + \frac{T - t}{T}A' + \frac{t}{T}b'
    \]
    is a $d$-dimensional Brownian bridge on $[0, T]$ with initial value $X_0' = A + A'$ and terminating value $X_T' = b + b'$.
    \item $\mu(t) := \E X_t = \frac{T - t}{T}\E A + \frac{t}{T}b$ for all $0 \leq t \leq T$.
    \item $k(s, t) := \Cov(X_s, X_t) = \frac{(T - t)(T - s)}{T^2}\Cov(A) + \left(s \land t  - \frac{s t}{T}\right) I_d$.
    \item For any $C > 0$, the process $(X_t')_{0 \leq t \leq CT}$ given by $X_t' := \sqrt{C}X_{t/C}$ is a $d$-dimensional Brownian bridge with initial point $\sqrt{C}A$ and terminal point $\sqrt{C}b$ on $[0, CT)$.
    \item If $A \sim \calN(\mu, \Sigma)$, then $(X_t)_{0 \leq t \leq T}$ is a continuous Gaussian process on $[0, T],$ and hence it's law is uniquely determined by $\mu$ and $k$.
\end{enumerate}
    
\end{lemma}

If $(B_t)_{t \geq 0}$ is a $d$-dimensional Brownian motion and $\tau$ is a stopping time with respect to the natural filtration $(\calF_t)_{t \geq 0}$, the strong Markov property for Brownian motion (see Theorem 2.20 of \citet{le2016brownian}) tells us that the process $(B_{\tau + t} - B_\tau)_{t \geq 0}$ is also a $d$-dimensional Brownian motion that is independent of $\calF_{\tau}$. While we need to be a little more careful with scaling in the setting of Brownian bridges, we can show a similar strong Markov property.

\begin{lemma}
\label{lem:strong_markov}
Let $(X_t)_{0 \leq t \leq 1}$ be a standard $d$-dimensional Brownian bridge with $X_0 = A$ and $X_1 = b$, and let $(\calG_t)_{0 \leq t \leq 1}$ be the corresponding natural filtration. Let $\tau$ be a $(\calG_t)$ stopping time. Let $(X_t^{(\tau)})_{0 \leq t \leq 1 - \tau}$ be the process defined by $X_t^{(\tau)} := X_{t + \tau} - \frac{1 - \tau - t}{1 - \tau}X_\tau - \frac{t}{1 - \tau}b$, and define the rescaled process $(Y_t^{(\tau)})_{0 \leq t \leq 1}$ by
\[
Y_t^{(\tau)} := \sqrt{1 - \tau}X^{(\tau)}_{t/(1 - \tau)}.
\] 
Then, $(Y_t)_{0 \leq t \leq 1}$ is a standard Brownian bridge with $Y_0 = Y_1 = 0$ independent of $\calG_\tau$.
\end{lemma}

\begin{proof}
\textbf{Step 1: reduction to the case $a = b = 0$:}
First, we note that it suffices to prove the result when $A = a$ is a constant. If we prove the result in this case, we note we have by the tower rule for conditional expectations that, for any event $E$,
\[
\P(Y^{(\tau)} \in E) = \E\left[\P\left(Y^{(\tau)} \in E \mid A \right)\right] = \E\left[\P\left(Z \in E\right)\right] = \P\left(Z \in E \right),
\]

where $(Z_t)_{0 \leq t \leq 1}$ is a Brownian bridge with $Z_0 = Z_1 = 0$. Next, note it suffices to prove the result in the case $a = b = 0.$ Let $(X_t)_{0 \leq t \leq 1}$ be a Brownian bridge satisfying $X_0 = a$ and $X_1 = b$. Define another process $(X_t')_{t \geq 0}$ on the same probability space by $X_t' := X_t - (1 - t)a - tb$. By the first part of Lemma~\ref{lem:bb_prop}, $(X_t')_{t \geq 0}$ is a Brownian bridge on $[0, 1]$ with initial point $X_0' = 0$ and $X_1' = 0$. Clearly, the natural filtration $(\calG_t)_{0 \leq t \leq 1}$ for $(X_t)_{0 \leq t \leq 1}$ is also the natural filtration for $(X_t')_{0 \leq t \leq 1}$. Further, a simple calculation yields that for any fixed $0 \leq s \leq t \leq 1$, $X_t^{(s)'} = X_t^{(s)}$. Thus it also follows that $Y_t^{(\tau)'} = Y_t^{(\tau)}$ for all $(\calG_t)$ stopping times $\tau$ and all $0 \leq t \leq 1$.

\textbf{Step 2: considering when $\tau = T$ is deterministic:}
Thus, going forward we consider the case where $(X_t)_{0 \leq t \leq 1}$ is a Brownian bridge with $X_0 = X_1 = 0$. Clearly it suffices to consider $(X_t)_{0 \leq t \leq 1}$ to be one-dimensional in what follows, as in the multivariate case the coordinates of $X$ are independent one-dimensional Brownian bridges. We first consider the case where $\tau = T$ is a constant time. In this case, the process $(Z_t)_{0 \leq t \leq 1}$ given by
\[
Z_t := \begin{cases} X_t \qquad \text{for } 0 \leq t < T, \\
X_{t - T}^{(T)} \qquad \text{for } T \leq t \leq 1   
\end{cases}
\]
is clearly a Gaussian process on $[0, 1]$ that is continuous on $[0, T)$ and $[T, 1]$. To show the result, we must show (1) for any $s \in [0, T), t \in [T, 1]$, $k(s, t) := \Cov(Z_s, Z_t) = 0$ (this implies $X^{(T)}$, and hence $Y^{(T)}$ is independent of $\calG_T$), (2) $\mu(t) := \E Z_t = 0$ for all $t \in [T, 1]$, and (3) $k(s, t) := \Cov(Z_s, Z_t)= (s - T)\land (t - T) - \frac{(s - T)(t - T)}{1 - T}$ for all $s, t \in [T, 1]$ (these final two points show the law of $X^{(T)}$ is that of a Brownian bridge since we already have sample path continuity).

We now check each of these properties. In what follows, recall that $X_t = B_t - tB_1$ for some (now one-dimensional) Brownian motion $(B_t)_{0 \leq t \leq 1}$, and remember that $\Cov(B_s, B_t) = s\land t$. 
\begin{enumerate}
    \item For $s \in [0, T)$ and $t \in [0, 1 - T]$, we have (assuming for now that $\E[X_t^{(T)}] = 0,$ which we confirm in a later point)
    \begin{align*}
    \Cov(X_s, X_t^{(T)}) &= \E\left[X_s\left(X_{t + T} - \frac{1 - T - t}{1 - T}X_T\right)\right] \\
    &= \E\left[X_s X_{t + T}\right] - \frac{1 - T - t}{1 - T}\E\left[X_s X_t\right]
    &= s(1 - t - T) + \frac{1 - T - t}{1 - T}s(1 - T) \\
    &= 0,
    \end{align*}
    which confirms the first point.
    \item For any $t \in [0, 1 - T]$, we have
    \begin{align*}
        \E\left[X_t^{(T)}\right] &= \E\left[B_{t + T} - (t + T)B_1 - \frac{1 - T - t}{1 - T}B_{T} + \frac{1 - T - t}{1 - T}TB_1\right] = 0,
    \end{align*}
    proving the second point.
    \item Lastly, using property 3 of Lemma~\ref{lem:bb_prop}, for $s, t \in [0, 1 - T]$ s.t.\ $s < t$, we have
    \begin{align*}
    \Cov\left(X_s^{(T)}, X_t^{(T)}\right) &= \E\left[\left(X_{s + T} - \frac{1 - T - s}{1 - T}X_T\right)\left(X_{t + T} - \frac{1 - T - t}{1 - T}X_T\right)\right] \\
    &= \left\{(s + T) - (s + T)(t + T)\right\} - \frac{1 - T - t}{1 - T}\left\{T - (s + T)T\right\} \\
    &\qquad - \frac{1 - T - s}{1 - T}\left\{T - (t + T)T\right\} + \frac{(1 - T - t)(1 - T - s)}{(1 - T)^2}\left\{T - T^2\right\} \\
    &= \frac{1}{1 - T}\Big[(s + T)(1 - T - t)(1 - T) - T(1 - T - s)(1 - T - t)\Big]\\
    &= \frac{(1 - T - t)s}{(1 - T)} = s - \frac{st}{1 - T}.
    \end{align*}
\end{enumerate}

Since we have shown that, for any $T \in [0, 1],$ $Y^{(T)}$ is independent of $\calG_T$, we have that, for any $E \in \calG_t$ and any bounded, any fixed times $0 \leq t_1 < t_2 < \cdots < t_p \leq 1$, and continuous function $F : \R^{dp} \rightarrow \R_{\geq 0},$

\[
\E\mathbbm{1}_E F(Y_{t_1}^{(T)}, \dots, F_{t_p}^{(T)}) = \P(A) \E F(X_{t_1}, \dots, X_{t_p}),
\]
which is a fact we will use in the sequel.

\textbf{Step 3: generalizing to general stopping times:}

We now emulate a standard proof of the strong Markov property for Brownian motion to extend to the case where $\tau$ is a $(\calG_t)_{0 \leq t \leq 1}$ stopping time (in particular, the proof of Theorem 2.20 in \citet{le2016brownian}).

It suffices to show that, for any $A \in \calG_\tau$, $0 \leq t_1 < t_2 < \cdots < t_p \leq 1$, and $F : \R^{dp} \rightarrow \R_{\geq 0}$ continuous and bounded that
\[
\E\mathbbm{1}_AF(Y_{t_1}^{(\tau)}, \dots, Y_{t_p}^{(\tau)}) = \P(A)\E F(X_{t_1}, \dots, X_{t_p}).
\]

As noted in \citet{le2016brownian}, this not only proves the independence of $(Y_t^{(\tau)})$ and $\calG_\tau$, but also demonstrates by taking $A = \Omega$ (where $(\Omega, \calF, \P)$ is the underlying probability space) that $(Y_t^{(\tau)})$ and $(X_t)$ have the same finite-dimensional distributions, and hence $(Y_t^{(\tau)})$ is a standard $d$-dimensional Brownian bridge since sample paths are continuous.

For $n$ a positive integer and $T \in \R$, define $T\vert_n := \min\{ k2^{-n} : k \in \Z,\; k2^{-n} \geq T\}$, i.e.\ $T\vert_n$ is the smallest real of the form $k2^{-n}$ that is greater than or equal to $T$. A straightforward expansion of $Y_t^{(\tau\vert_n)}$ yields that, for any $t \in [0, 1]$, we have $Y_t^{(\tau\vert_n)} \xrightarrow[n \rightarrow \infty]{} Y_t^{(\tau)},$ and thus bounded convergence yields
\begin{align*}
\E\mathbbm{1}_A F(Y_{t_1}^{(\tau)}, \dots, Y_{t_p}^{(\tau)}) &= \lim_{n \rightarrow \infty}\E \mathbbm{1}_A F(Y_{t_1}^{(\tau\vert_n)}, \dots, Y_{t_p}^{(\tau\vert_n)}) \\
&= \lim_{n \rightarrow \infty}\sum_{k = 0}^{2^n}\E\mathbbm{1}_A\mathbbm{1}_{E_n^k}F(Y_{t_1}^{(\tau\vert_n)}, \dots, Y_{t_p}^{(\tau\vert_n)}) \\
&= \lim_{n \rightarrow \infty}\sum_{k = 1}^{2^n}\E\mathbbm{1}_A\mathbbm{1}_{E_n^k} F(Y_{t_1}^{(k2^{-n})}, \dots, Y_{t_p}^{(k2^{-n})}) \\
&= \lim_{n \rightarrow \infty}\sum_{k = 1}^{2^n}\P(A \cap E_n^k) \E F(Y_{t_1}^{(k2^{-n})}, \dots, Y_{t_p}^{(k2^{-n})})\\
&= \P(A)\E F(X_{t_1}, \dots, X_{t_p}),
\end{align*}
where $(X_t)_{0 \leq t \leq 1}$ is a standard $d$-dimensional Brownian bridge, proving the desired result. In the above, $E_n^k := \{(k - 1)2^{-n} < \tau \leq k2^{-n}\}$, and we use the identity $\mathbbm{1}_{E_n^k}F(Y_{t_1}^{(\tau\vert_n)}, \dots, Y_{t_p}^{(\tau\vert_n)}) = \mathbbm{1}_{E_n^k} F(Y_{t_1}^{(k2^{-n})}, \dots, Y_{t_p}^{(k2^{-n})})$. The second to last inequality follows from applying the result where $t$ is a deterministic time, noting that the event $A \cap E_n^k$ is $\calG_{k2^{-n}}$-measurable.

Thus, we have shown the desired result.
\end{proof}

\begin{corollary}
\label{cor:strong_markov}
Let $(X_t)_{0 \leq t \leq 1}$ be a $d$-dimensional Brownian bridge with $X_0 = A$ and $X_1 = b$, where $A$ is a random variable. Let $(\calG_t)_{0 \leq t \leq 1}$ be the corresponding natural filtration. Let $\tau$ be a $(\calG_t)$ stopping time. Then, for any $G \in \calG_\tau$, the conditional law of the process $(X_t)_{\tau \leq t \leq 1}$ given $\{\tau = T, X_\tau = x\} \cap G$ is that of a Brownian bridge on $[T, 1]$ with initial value $X_T = x$ and terminal value $X_1 = b$, i.e.
\[
\P(X \in \cdot \mid \tau = T, X_\tau = x, G) = \P(S \in \cdot),
\]
where $(S_t)_{T \leq t \leq 1}$ is a Brownian bridge on $[T, 1]$ with $S_T = x$ and $S_1 = b$.
\end{corollary}

\begin{proof}
    Let $(Z_t)_{0 \leq t \leq 1}$ be a Brownian bridge wth $Z_0 = Z_1 = 0$. Applying the tower rule for conditional expectation alongside Lemma~\ref{lem:strong_markov} gives us, for all $E \in \calF$,
    \[
    \P\left(Y^{(\tau)} \in E \mid \tau, X_\tau, \mathbbm{1}_G\right) = \E\left[\P\left(Y^{(\tau)} \in E \mid \calF_\tau\right)\right] = \P(Z \in E).
    \]
    Thus, with probability one over the joint distribution of $(X_\tau, \tau, \mathbbm{1}_G)$, we have
    \[
    \P\left(Y^{(\tau)} \in E \mid X_\tau = x, \tau = t, G\right) = \P(Z \in E).
    \]
    With Lemma~\ref{lem:bb_prop}, we know that, since $Y^{(\tau)}$ is a Brownian bridge with $Y^{(\tau)}_0 = Y^{(\tau)}_1 = 0$ on this event, then, $\frac{1}{\sqrt{1 - T}}Y^{(\tau)}_{t(1 - T)} = X_{t + T} - \frac{1 - T - t}{1 - T} x  + \frac{t}{1 - T}b$ is a Brownian bridge with initial and terminal value $0$ on $[0, 1 - T]$. The remainder of the result follows by adding $\frac{1 - T - t}{1 - T} x  - \frac{t}{1 - T}b$, applying the first part of Lemma~\ref{lem:bb_prop}, and reindexing the process to be defined on $[T, 1]$.
\end{proof}

Lemma~\ref{lem:strong_markov} and Corollary~\ref{cor:strong_markov} above show that the conditional distributions of Brownian bridges, even at stopping times, are very well-behaved --- the conditional distributions are exactly that of another Brownian bridge. We aim to apply these results to our analysis of the privacy loss of the Brownian mechanism as follows. We will shortly that the distribution of the outputs of the Brownian mechanism, which can be viewed as a Brownian motion being run in reverse, can be equivalently viewed as a Brownian bridge with random (particularly, multivariate Gaussian) initial state and fixed terminating state. Coupling this with the above strong Markov property, we will show that even when an analyst picks arbitrarily complicated stopping functions, the privacy loss looks as if the inspection times were fixed in advance.

First, we show that, for a fixed number $n$ of time functions, the privacy loss is exactly as outlined in the statement of Proposition~\ref{prop:target_ploss}.

\begin{lemma}
\label{lem:ploss_fix}
Let $n \in \N$ be arbitrary, and let $T_1, \dots, T_n$ be decreasing (i.e.\ non-increasing) time functions. Let $p^\nu_{1:n}$ denote the joint density of $(B^\nu_{T_1^\nu}, \dots, B^\nu_{T_n^\nu})$, where $(B^\nu_t)_{t \geq 0}$ is a $d$-dimensional Brownian motion starting at $\nu \in \R^d$ and $T_m^\nu := T_m\left(B_{T_1^\nu}^\nu, \dots, B_{T_{m - 1}^\nu}^\nu\right)$. Then, for any $y_1, \dots, y_n \in \R^d$, we have\footnote{Since we may have $T_m = T_{m - 1}$ for some $m$, we adopt the convention that when $y_m = y_{m - 1}$, $\exp\left(\frac{-(y_{m} - y_{m - 1})^2}{2(T_{m} - T_{m - 1})}\right) = 1$. Likewise, when $y_m \neq y_{m - 1}$ in this setting, we adopt $\exp\left(\frac{-(y_m - y_{m - 1})^2}{2(T_m - T_{m - 1})}\right) = 0$. After the proof of this lemma, only the former case will occur.}
\[
p^\nu_{1:n}(y_1, \dots, y_n) \propto_\nu \exp\left(-\frac{\|y_{n} - \nu\|^2}{2T_n}\right) \prod_{m = 2}^n \exp\left(\frac{-\|y_{m - 1} - y_m\|^2}{2(T_{m - 1} - T_m)}\right),
\]
where $T_m = T_m(y_1, \dots, y_{m - 1})$ for notational convenience and $\propto_\nu$ indicates that the constant of proportionality does not depend on $\nu$.

\end{lemma}

\begin{proof}
We prove the result by induction on $n$, with the base case of $n = 1$ being trivial. Assume now the result holds for $n$. Recall that the first time function $T_1$ is simply a constant. Define the ``backwards'' process $(X_t^\nu)_{0 \leq t \leq T_1}$ by $X_t^\nu := B_{T_1 - t}^\nu$, and let $(\calG_t)_{0 \leq t \leq T_1}$ be the corresponding natural filtration, i.e.\ $\calG_t := \sigma(X_s^\nu : s \leq t) = \sigma(B^\nu_{T_1 - s} : s \leq t)$. Inspection yields that $(X_t^\nu)_{0 \leq t \leq T_1}$ is a Brownian bridge with $X_{0}^\nu \sim \calN(\nu, T_1 I_d)$ and $X_{T_1}^\nu = \nu$.

First, we note that the strong Markov property (in particular Corollary~\ref{cor:strong_markov}) yields that, for any $(\calG_t)$ stopping times $\tau_1 \leq \dots \leq \tau_n$, the law of of $(X_t^\nu)_{\tau_n \leq 1 \leq T_1}$ conditional on the event $\{X_{\tau_1}^\nu = y_1, \dots, X_{\tau_n}^\nu = y_n\}$ is that of a Brownian bridge with initial point $X_{\tau_n}^\nu = y_n$ and terminal point $X_{T_1}^\nu = \nu$. Applying this in the case $\tau_m = T_1 - T_m$, this yields that the conditional law of $(X_t^\nu)_{T_1 - T_n \leq t \leq T_1}$ given $\{X_{0}^\nu = y_1, X_{T_1 - T_2}^\nu = y_2, \dots, X_{T_1 - T_n}^\nu = y_n\}$ is a Brownian bridge with initial point $X_{T_1 - T_n}^\nu = y_n$ and terminal point $X_{T_1}^\nu = \nu$.  But, this is equivalent to saying the conditional law of $(B_t^\nu)_{0 \leq t \leq T_n}$ given $\{B_{T_n}^\nu = y_n, \dots, B_{T_1}^\nu = y_1\}$ is that of a Brownian bridge with initial value $B_{0}^\nu = \nu$ and terminal value $B_{T_n}^\nu = y_n$.

Next, note that, on the event $\{B_{T_n}^\nu = y_n, \dots, B_{T_1}^\nu = y_1\}$, the time function $T_{n + 1} = T_{n + 1}(y_1, \dots, y_n)$ is constant in value. Following the from the preceding paragraph, the conditional density $p^\nu_{1:n + 1}(y_{n + 1} \mid y_1, \dots, y_n)$ is just that of a Brownian bridge with initial value $B_0^\nu = \nu$ and $B_{T_n}^\nu = y_n$ inspected at time $T_{n + 1}$. That is, from using the covariance and mean expressions for a Brownian bridge along with the fact it is a Gaussian process, we have by Lemma~\ref{lem:bb_prop}
\[
p_{1:n + 1}^\nu(y_{n + 1} \mid y_1, \dots, y_n) \propto_\nu \exp\left(-\frac{\left\|y_{n + 1} - \nu - \frac{T_{n + 1}}{T_n}(y_{n} - \nu)\right\|^2}{2(T_{n} - T_{n + 1})}\cdot\frac{T_{n + 1}}{T_n}\right).
\]

Thus, applying Bayes rule for densities alongside the inductive hypothesis, we have

\begin{align*}
&p^\nu_{1:n + 1}(y_1, \dots, y_{n + 1}) = p^\nu_{1 : n}(y_1, \dots, y_n)p^\nu_{1:n + 1}(y_{n + 1} \mid y_1, \dots, y_n) \\
&\qquad \propto_\nu \exp\left(-\frac{\|y_n - \nu\|^2}{2T_n}\right)\cdot \left(\prod_{m = 2}^n \exp\left(\frac{-\|y_{m - 1} - y_m\|^2}{2(T_{m - 1} - T_m)}\right)\right)\cdot\exp\left(-\frac{\left\|y_{n + 1} - \nu - \frac{T_{n + 1}}{T_n}(y_n - \nu)\right\|^2}{2(T_n - T_{n + 1})}\cdot\frac{T_{n + 1}}{T_n}\right) \\
&\qquad = \exp\left(-\frac{\|y_{n + 1} - \nu\|^2}{2T_{n + 1}}\right)\cdot\prod_{m = 2}^{n + 1} \exp\left(\frac{-\|y_{m - 1} - y_m\|^2}{2(T_{m - 1} - T_m)}\right),
\end{align*}
which proves the desired claim.
\end{proof}

With the above lemma, which shows that Proposition~\ref{prop:target_ploss} holds when the number of time functions is constant, we can now prove that Proposition~\ref{prop:target_ploss} holds in full generality. The idea behind the general proof is as follows. First, we consider the setting where an analyst has a sequence of time functions $T_1, T_2, \dots$ and uses a stopping function $N$ that satisfies $N((y_n)_{n \geq 1}) \leq n$ for all possible strings of inputs. We then construct a sequence of exactly $n$ time functions $S_1, \dots, S_n$ such that $p_{1:n}^\mu\left(B^{\nu}_{S_1^\nu}, \dots, B_{S_n^\nu}^\nu\right) = p_{1:N}^\mu\left(B^\nu_{T_1^\nu}, \dots, B^\nu_{T_{N^\nu}^\nu}\right)$. Then, in the general case where we only assume $N((y_n)_{n \geq 1}) < \infty$ for all sequences $(y_n)_{n \geq 1}$, for any $\delta > 0, \nu \in \R^d$, there is some $n_\delta^\nu$ such that $\P\left(N\left(\left(B_{T_n^\nu}^\nu\right)_{n \geq 1}\right) \leq n_{\nu, \delta}\right) \geq  1 - \delta$, which will allow us to apply our argument from the setting where $N$ is bounded alongside a limiting argument.

With the above brief description of our technique at hand, we now prove Proposition~\ref{prop:target_ploss}.

\begin{proof}[Proof of Proposition~\ref{prop:target_ploss}]

By assumption, for all sequences $(y_m)_{m \geq 1}$ of elements of $\R^d$, we have $N\left((y_m)_{m \geq 1}\right) \leq n$ for some fixed natural number $n \in \N$. If $(T_m)_{m \geq 1}$ is the original sequence of stopping functions, define a new sequence by $S_m := T_{m \land N}$ for all $m \in [n]$.\footnote{While $N$ technically accepts an infinite sequence $(y_n)_{n \geq 1}$ of vectors as input, by definition, checking $N\left((y_n)\right) \leq m$ only requires examining the first $m$ elements of the sequence $y_1, \dots, y_m$.} 

It is straightforward to see that, for any $\mu, \nu \in \R^d$, $p^\mu_{1:N}\left(B_{T_1^\nu}^\nu, \dots, B_{T_{N^\nu}^\nu}^\nu\right) = p^\mu_{1:n}\left(B^\nu_{T_{1 \land N^\nu}}, \dots, B^\nu_{T_{n \land N^\nu}^\nu}\right) = p^\mu_{1:n}\left(B_{S_1^\nu}^\nu, \dots, B_{S_n^\nu}^\nu\right)$. Moreover, Lemma~\ref{lem:ploss_fix} yields that
\begin{align*}
\frac{p^\nu_{1:n}\left(B_{S_1^\nu}^\nu, \dots, B_{S_n^\nu}^\nu\right)}{p^\mu_{1:n}\left(B_{S_1^\nu}^\nu, \dots, B_{S_n^\nu}^\nu\right)} = \frac{\exp\left(-\frac{\left\|B_{S_n^\nu}^\nu - \nu\right\|^2}{2S_n^\nu}\right)}{\exp\left(-\frac{\left\|B_{S_n^\nu}^\nu - \mu\right\|^2}{2S_n^\nu}\right)} = \frac{\exp\left(-\frac{\left\|B_{T_{N^\nu}^\nu}^\nu - \nu\right\|^2}{2T_{N^\nu}^\nu}\right)}{\exp\left(-\frac{\left\|B_{T_{N^\nu}^\nu}^\nu - \mu\right\|^2}{2T_{N^\nu}^\nu}\right)},
\end{align*}
which is just the ratio between the density of a  $\calN\left(\nu, T_{N^\nu}^\nu\right)$ random variable and a $\calN\left(\mu, T_{N^\nu}^\nu\right)$ random variable evaluated at $B_{T_{N^\nu}^\nu}^\nu$, proving the desired result.
\end{proof}

We now prove Theorem~\ref{thm:bm_ploss}, which gives a closed form characterization of the Brownian mechanism. In what follows, we use the same notation for the density of Brownian motion as in the above proof.

\begin{proof}[\textbf{Proof of Theorem~\ref{thm:bm_ploss}}]
The second statement of the theorem is trivial and follows from our assumption of bounded $\ell_2$ sensitivity. Hence, we only prove the first statement below.

From the results of Lemma~\ref{lem:bm_nr}, we have
\begin{align*}
\calL_{1:N(x)}^{\BM}(x, x') &= \log\left(\frac{p_{T_{N(x)}(x)}^{f(x)}(\BM_n(x))}{p_{T_{N(x)}(x)}^{f(x')}(\BM_n(x))}\right) \\
&=- \frac{1}{2} \left[  \frac{|| B_{T_{N(x)}} -f(x)||_2^2}{T_{N(x)}(x)}  - \frac{|| B_{T_{N(x)}} -f(x')||_2^2}{T_{N(x)}(x)} \right] 
\end{align*}

Without loss of generality, and for the sake of simplicity, $f(x) = 0$.  The privacy loss can be written as 
\begin{align*}
\calL_{1:N(x)}^{\BM}(x, x') &=  \frac{1}{2T_{N(x)}(x)}\left( - ||B_{T_{N(x)}(x)}||_2^2 + ||B_{T_{N(x)}(x)} - f(x')||_2^2\right) \\
&= -\frac{1}{T_n(x)}\langle B_{T_{N(x)}(x)}, f(x') \rangle + \frac{1}{2T_{N(x)}(x)}||f(x')||_2^2 \\
&= -\frac{||f(x')||_2}{T_{N(x)}(x)}\left\langle B_{T_{N(x)}(x)}, \frac{f(x')}{||f(x')||_2}\right\rangle + \frac{1}{2T_{N(x)}(x)}||f(x')||_2^2 \\
&= -\frac{||f(x')||_2}{T_{N(x)}(x)}\left\langle B_{T_{N(x)}(x)}, \frac{f(x')}{||f(x')||_2}\right\rangle + \frac{1}{2T_{N(x)}(x)}||f(x')||_2^2 \\
&= -\frac{||f(x')||_2}{T_{N(x)}(x)}W_{T_n(x)} + \frac{1}{2T_{N(x)}(x)}||f(x')||_2^2.
\end{align*}
Note that the last inequality follows from the fact that if $(B_t)_{t \geq 0}$ is a d-dimensional Brownian motion and $z \in \R^d$ is a unit vector under the $\ell_2$ norm, then the process $W_t := \langle z, B_t \rangle$ is a standard Brownian motion. Noting that $(-W_t)_{t \geq 0}$ is also a Brownian motion furnishes the result.
\end{proof}

We now use the characterization of privacy loss in Theorem~\ref{thm:bm_ploss} alongside the time-uniform concentration results for continuous time martingales found in Appendix~\ref{app:martingale} to construct two general families of privacy boundaries. We now prove Theorem~\ref{thm:priv_bound}.

\begin{proof}[\textbf{Proof of Theorem~\ref{thm:priv_bound}}]

Recall from Theorem~\ref{thm:bm_ploss} that we have the following bound
\[
\calL_{1:N(x)}^{\BM}(x, x') \leq \frac{\Delta^2}{2T_{N(x)}(x)} + \frac{\Delta}{T_{N(x)}(x)}W_{T_{N(x)}(x)}^+,
\]
where $A^+ :=\max(A, 0)$. First, by leveraging Lemma~\ref{lem:mixture}, we see that, with probability at least $1 - \delta$, we have
\[
\calL_{1:N(x)}^{\BM}(x, x') \leq \frac{\Delta^2}{2T_{N(x)}(x)} + \frac{\Delta}{T_{N(x)}(x)}\sqrt{2(T_{N(x)}(x) + \rho)\log\left(\frac{1}{\delta}\sqrt{\frac{T_{N(x)}(x) + \rho}{\rho}}\right)} = \psi_\rho^M(T_{N(x)}(x)),
\]
proving that $\psi_\rho^M$ is a valid $\delta$-privacy boundary. Likewise, by Lemma~\ref{lem:line_cross}, we have that
\[
\calL_{1:N(x)}^{\BM}(x, x') \leq \frac{\Delta^2}{2T_{N(x)}(x)} + \frac{\Delta}{T_{N(x)}(x)}(a T_{N(x)}(x) + b) = \frac{\Delta}{T_{N(x)}(x)}\left(\frac{\Delta}{2} + b\right) + \Delta a = \psi^L_{a, b}(T_{N(x)}(x)),
\]
showing $\psi_{a, b}^L$ is a valid $\delta$-privacy boundary.

\end{proof}

%% file: appendices/rat_proof.tex
\section{Proofs From Section~\ref{sec:rat}}
\label{app:rat}

In this appendix, we provide proofs of the results in Section~\ref{sec:rat}. We start by proving the privacy guarantees for $\RAT$.

\begin{proof}[\textbf{Proof of Theorem~\ref{thm:rat}}]
For $\RAT$ as described in Algorithm~\ref{alg:rat}, on the event $\{N(x) = n\}$, all information leaked about the underlying private dataset is contained in $\Alg_{1:n}(x)$ and $\alpha_{1:n}(x)$, where $\alpha_n(x)$ is defined to be the $n$th bit output by $\RAT$. For any $y \in \calX$, let $q^{y}_{1:n}$ denote the joint density of $(\Alg_{1:n}(y), \alpha_{1:n}(y))$, $p^{y}_{1:n}$ the marginal density of $\Alg_{1:n}(y)$, and $p^y_{1:n}(\cdot \mid \cdot)$ the conditional pmf of $\alpha_{1:n}(y)$ given the observed values of $\Alg_{1:n}(y)$.  As such, for any neighboring datasets $x \sim x'$, on the event $\{N(x) = n\}$, the privacy loss of $\RAT$, denoted by $\calL^{\mathrm{RAT}}(x, x')$, is given by
\begin{align*}
    \calL^{\mathrm{RAT}}_{1:n}(x, x') &= \log\left(\frac{q^x_{1:n}(\Alg_{1:n}(x), \alpha_{1:n}(x))}{q^{x'}_{1:n}(\Alg_{1:n}(x), \alpha_{1:n}(x))}\right) \\
    &= \log\left(\frac{p^{x}_{1:n}(\Alg_{1:n}(x))}{p^{x'}_{1:n}(\Alg_{1:n}(x))}\right) + \log\left(\frac{p^{x}_{1:n}(\alpha_{1:n}(x) \mid \Alg_{1:n}(x))}{p^{x'}_{1:n}(\alpha_{1:n}(x) \mid \Alg_{1:n}(x))}\right) \\
    &= \log\left(\frac{p^{x}_{1:n}(\Alg_{1:n}(x))}{p^{x'}_{1:n}(\Alg_{1:n}(x))}\right) + \log\left(\frac{p^{x}_{1:n}(0^{n - 1}1 \mid \Alg_{1:n}(x))}{p^{x'}_{1:n}(0^{n - 1}1 \mid \Alg_{1:n}(x))}\right) \\
    &= \calL_{1:n}^{\Alg}(x, x') + L_n(x, x'),
\end{align*}
where $0^{n-1}1$ denotes the string of $n - 1$ 0's followed by a single 1. In the last line we leverage the definition of the privacy loss between $\Alg_{1:n}(x)$ and $\Alg_{1:n}(x')$ and define 
\[
L_n(x, x') := \log\left(\frac{p^{x}_{1:n}(0^{n - 1}1 \mid \Alg_{1:n}(x))}{p^{x'}_{1:n}(0^{n - 1} 1 \mid \Alg_{1:n}(x))}\right).
\]

Now, to finish the result, it suffices to prove that, for any $n$, $L_n(x, x') \leq \Epsilon_n(\Alg_{1:n - 1}(x))$. Without loss of generality, we can assume all thresholds take the same value $\tau$ across rounds, as we can always define the shifted function $u_n'(\Alg_{1:n}(x), x) := u_n(\Alg_{1:n}(x), x) - \tau_n + \tau$.
To prove our desired inequality, we proceed largely in the same way as the proof of $\AT$ found in \citet{lyu2016understanding}, noting that conditioning on $\Alg_{1:n}(x)$ serves to fix the utility functions $u_1(\Alg_1(x), \cdot), \dots, u_n(\Alg_{1:n}(x), \cdot)$ and the privacy levels $\Epsilon_1, \Epsilon_2(\Alg_1(x)), \dots, \Epsilon_n(\Alg_{1:n-1}(x))$. For simplicity, going forward, we refer to the former quantities as $u_1(\cdot), \dots, u_n(\cdot)$ and the latter quantities just as $\epsilon_1, \dots, \epsilon_n$. The only remaining caveat that we must take care in handling variable amount of noise on the threshold introduced by $\LNR$. Going forward, let $\P_{1:n}$ denote the conditional probability $\P(\cdot \mid \Alg_{1:n}(x))$. First, observe that we can write the numerator of $L_n(x, x')$ as
\begin{align*}
p^{x}\left(0^{n - 1}1 \mid \Alg_{1:n}(x)\right) = \int_{\R^n}g^\tau_{1:n}(s_1, \dots, s_n)\left(\prod_{i = 1}^{n - 1}\P_{1:n}\left(u_i(x) + \xi_i < s_i \right)\right)\P_{1:n}\left(u_n(x) + \xi_n \geq s_n \right)d\vec{s},
\end{align*}
where $g^\tau_{1:n}$ represents the density for the joint distribution of $(\tau + Z(2\Delta/\epsilon_m))_{m = 1}^n$, where $(Z(t))_{t \geq \eta}$ is as defined in Equation~\eqref{eq:lap_proc}.
We now need three inequalities. The first two are standard from the analysis of \citet{lyu2016understanding}, so we do not provide a proof. The third inequality is a product of our novel $\RAT$ mechanism, and hence we provide a proof. The inequalities are:
\begin{enumerate}
    \item For $i < n$ and fixed $s_i$, $\P_{1:n}(u_i(x) + \xi_i < s_i) \leq \P_{1:n}(u_i(x') + \xi_i < s_i + \Delta)$,
    \item for $i = n$ and any $s_n$, $\P_{1:n}(u_n(x) + \xi_n \geq s_n) \leq e^{\epsilon_n/2}\P_{1:n}(u_n(x') + \xi_n \geq s_n + \Delta)$, and
    \item for any $s_{1:n} \in \R^n$, $g^\tau_{1:n}(s_{1}, \dots, s_n) \leq e^{\epsilon_n/2}g^\tau_{1:n}(s_1 + \Delta, \dots, s_n + \Delta)$.
\end{enumerate}

We now prove the third inequality. We have that
\begin{align*}
    \frac{g^{\tau}_{1:n}(s_1, \dots, s_n)}{g^{\tau - \Delta}_{1:n}(s_1, \dots, s_n)} &= \frac{g^\tau_n(s_n)g^\tau_{1:n - 1}(s_1, \dots, s_{n - 1}\mid s_n)}{g^{\tau - \Delta}_n(s_n)g^{\tau - \Delta}_{1: n -1}(s_1, \dots, s_{n - 1}\mid s_n)} \\
    &= \frac{g^\tau_{n}(s_n)}{g^{\tau - \Delta}_n(s_n)} \leq e^{\epsilon_n/2},\end{align*}
where the first equality follows from applying Bayes rule to the joint densities of the noisy thresholds, and the second equality follows from the fact that $(Z(t))$ forms a Markov process. This in particular implies that the density conditional density given the $n$th threshold satisfies $g^a_{1:n -1}(s_1, \dots, s_{n - 1} \mid s_n) = g^b_{1: n -1}(s_1, \dots, s_{n - 1} \mid s_n)$ for all $a, b \in \R$. The last inequality follows from examining the ratio of densities of $\Lap(\tau, 2\Delta/\epsilon_n)$ and $\Lap(\tau - \Delta, 2\Delta/\epsilon_n)$ random variables. Now, observe that by a simple shift of parameters we have
\[
g^{\tau - \Delta}_{1:n}(s_1, \dots, s_n) = g^\tau_{1:n}(s_1 + \Delta, \dots, s_n + \Delta).
\]
Plugging this in, we have
\begin{align*}
&\;  p^{x}\left(0^{n - 1}1 \mid \Alg_{1:n}(x)\right) \\ &= \int_{\R^n}g^\tau_{1:n}(s_1, \dots, s_n)\left(\prod_{i = 1}^{n - 1}\P_{1:n}(u_i(x) + \xi_i < s_i)\right)\P_{1:n}(u_n(x) + \xi_n \geq s_n)d\vec{s} \\
   &\leq e^{\epsilon_n/2}\int_{\R^n}g^{T - \Delta}_{1:n}(s_1, \dots, s_n)\left(\prod_{i = 1}^{n - 1}\P_{1:n}(u_i(x) + \xi_i < s_i)\right)\P_{1:n}(u_n(x) + \xi_n \geq s_n)d\vec{s} \\
    &\leq e^{\epsilon_n}\int_{\R^n}g^{\tau- \Delta}_{1:n}(s_1, \dots, s_n)\left(\prod_{i = 1}^{n - 1}\P_{1:n}(u_i(x') + \xi_i < s_i + \Delta)\right)\P(u_n(x') + \xi_n \geq s_n + \Delta)d\vec{s} \\
    &= e^{\epsilon_n}\int_{\R^n}g^\tau_{1:n}(s_1, \dots, s_n )\left(\prod_{i = 1}^{n - 1}\P_{1:n}(u_i(x') + \xi_i < s_i)\right)\P_{1:n}(u_n(x') + \xi_n \geq s_n)d\vec{s} \\
    &= e^{\epsilon_n}p^{x'}\left(0^{n - 1}1 \mid \Alg_{1:n}(x)\right).
\end{align*}
Rearranging furnishes the desired result.
\end{proof}

We can also prove a corresponding utility guarantee for $\RAT$. As mentioned earlier, this utility guarantee is naive in the sense that it is derived from a union bound. Thus, instead of plotting the utility guarantee in our experiments in Section~\ref{sec:exp}, we instead plot empirically observed loss/accuracy. Additionally, for the utility guarantee to hold, the sequence of privacy functions $(\Epsilon_n)_{n \geq 1}$ must be constant functions, i.e. $\Epsilon_n = \epsilon_n$ for each $n$. We now state the formal, high-probability utility guarantee in the following proposition.

\begin{prop}
\label{prop:rat_utility}
Let $(p_n)_{n \geq 1}$ be a sequence of non-negative numbers such that $\sum_{i = 1}^\infty p_i = 1$, and let $\gamma \in (0, 1)$ be a confidence parameter. Define the sequence of parameters $(\eta_n)_{n \geq 1}$ by 
\[
\eta_n := \frac{4\Delta}{\epsilon_n}\left(\log\left(\frac{2}{\gamma}\right) - \log(p_n)\right).
\]
Then, if $N(x)$ is the time defined in Theorem~\ref{thm:rat}, with probability at least $1 - \gamma$, we have
\[
u_{N(x)}(x) \geq \tau_{N(x)} - \eta_{N(x)}.
\]
\end{prop}

\begin{proof}
The above utility guarantee follows from applying two simple union bounds. First, we have 
\[
\P\left(\bigcup_{n \geq 1}\{|\xi_n| \geq \eta_n/2\}\right) \leq \sum_{n \geq 1}\P(|\xi_n| \geq \eta_n/2) = \sum_{n \geq 1}\exp\left(\frac{-\epsilon_n \eta_n}{4\Delta}\right) = \frac{\gamma}{2}\sum_{n \geq 1}p_n = 1.
\]
Second, we have that
\[
\P\left(\bigcup_{n \geq 1}\{|\zeta_n| \geq \eta_n/2\}\right) \leq \sum_{n \geq 1}\P(|\zeta_n| \geq \eta_n/2) = \sum_{n \geq 1}\exp\left(\frac{-\epsilon_n \eta_n}{2\Delta}\right) \leq \frac{\gamma}{2}\sum_{n \geq 1}p_n = 1.
\]
Thus, with probability at least $1 - \gamma$, we have simultaneously for all $n \geq 1$ that $|\xi_n| \leq \eta_n/2$ and $|\zeta_n| \leq \eta_n/2$. Thus, with the same probability, on round $N(x)$, we have 
\[
u_{N(x)}(x) \geq \tau_{N(x)} - \eta_{N(x)}.
\]
\end{proof}

%% file: appendices/additional_NR.tex
\section{Proofs From Section~\ref{sec:lnr}}
\label{app:nr}
We first prove that the process defined in Equation~\eqref{eq:lap_proc} has Laplace marginal distributions. 

\begin{theorem}
Let $(Z_t)_{t \geq \eta}$ be the process defined in Equation~\eqref{eq:lap_proc}. Then, for any $t \geq \eta$, we have
\[
Z_t \sim \mathrm{Lap}(t).
\]
\end{theorem}
In what follows, we sometimes use the notation $Z(t)$ interchangeably with $Z_t$ for convenience.
\begin{proof}
Recall that if $X \sim \mathrm{Lap}(s)$, then $X$ has characteristic function $\varphi_s$ given by
\[
\varphi_s(\lambda) = \frac{1}{1 + \lambda^2 s^2}.
\]
Let $\phi$ denote the characteristic function of $Z_t - Z_\eta$. Since $Z_\eta$ and $Z_t - Z_\eta$ are independent, to show $Z_t \sim \mathrm{Lap}(t)$, it suffices to show that
\[
\phi(\lambda) = \frac{\varphi_t(\lambda)}{\varphi_\eta(\lambda)} = \frac{1 + \lambda^2 \eta^2}{1 + \lambda^2 t^2}.
\]

Now, observe that the inhomogenous Poisson process $(P_t)_{t \geq \eta}$ can be written as $(\widetilde{P}(e^{t/2}))_{t \geq \log(\eta^2)}$ where $\widetilde{P}$ is a homogeneous Poisson process with rate $\lambda = 1$ on $[\log(\eta^2), \infty)$. In terms of the process $\widetilde{P}$, we can consider the process $(\widetilde{Z}_t)_{t \geq \log(\eta^2)}$ given by
\[
\widetilde{Z}_t = \sum_{n  \leq \widetilde{P}_t}\mathrm{Lap}\left(e^{\widetilde{\calT}_n/2}\right),
\]
where $\widetilde{\calT}_n := \inf\{t \geq \log(\eta^2) : \widetilde{P}_t \geq n\}$ and $\widetilde{\calT}_0 = \log(\eta^2)$. It is easy to see that 
\[\widetilde{Z}(\log(t^2)) - \widetilde{Z}(\log(\eta^2)) =_d Z_t - Z_\eta.\]
Leveraging this identity, it follows that we have
\begin{align}
    \phi(\lambda) &= \E\left[e^{i \lambda (Z_t - Z_\eta)}\right] = \E\left[e^{i \lambda (\widetilde{Z}(\log(t^2)) - \widetilde{Z}(\log(\eta^2)))}\right] \nonumber \\
    &= \sum_{n = 0}^\infty \frac{\eta^2}{t^2}\frac{\left[\log(t^2/\eta^2)\right]^n}{n!}\int_{\log(\eta^2) \leq u_1 < u_2 < \cdots < u_n \leq \log(t^2)}f^{(n)}(u_1, \dots, u_n)\prod_{i = 1}^n\E\left[e^{i\lambda \mathrm{Lap}(e^{u_i/2})}\right]d\mathbf{u} \nonumber \\
    &= \frac{\eta^2}{t^2}\sum_{n = 0}^\infty \int_{\log(\eta^2) \leq u_1 < u_2 < \cdots < u_n \leq \log(t^2)}\prod_{i = 1}^n\frac{1}{1 + \lambda^2e^{u_i}}d\mathbf{u} \label{eq:char_fn}.
\end{align}

In the above, $f^{(n)}(u_1, \dots, u_n) := \frac{n!}{[\log(t^2/\eta^2)]^n}$ is the distribution of the order statistics $(U_{(1)}, \dots, U_{(n)})$ of $n$ i.i.d. random variables that are uniform on $[\log(\eta^2), \log(t^2)]$. Essentially, what we have done is \textit{first} conditioned of the number of Poisson arrivals that occur in the interval $[\log(\eta^2), \log(t^2)]$. Then, on the event $\{N(t) = n\}$, we condition again on the location of the $n$ arrivals, which we know to be uniformly distributed across the time interval. Once the arrival locations are known, we can compute the conditional characteristic function, which is the the product of characteristic functions as illustrated in the integral above.  

Now, we show inductively that
\[
\int_{\log(\eta^2) \leq u_1 < u_2 < \cdots < u_n \leq \log(t^2)}\prod_{i = 1}^n\frac{1}{1 + \lambda^2e^{u_i}}d\mathbf{u} = \frac{1}{n!}\left[\log\left(\frac{t^2}{\eta^2}\frac{1 + \lambda^2 \eta^2}{1 + \lambda^2 t^2}\right)\right]^n.
\]

The base case of $n = 1$ is trivially true. Now, we have that
\begin{align*}
&\int_{\log(\eta^2) \leq u_1 < u_2 < \cdots < u_n \leq \log(t^2)}\prod_{i = 1}^n\frac{1}{1 + \lambda^2e^{u_i}}d\mathbf{u} \\
&= \int_{u_1 = \log(\eta^2)}^{\log(t^2)}\frac{1}{1 + \lambda^2e^{u_1}}\int_{u_1 < u_2 < \dots < u_n}\prod_{i = 2}^n\frac{1}{1 + \lambda^2 e^{u_i}}d\mathbf{u}_{-1}du_1 \\
&= \frac{1}{(n - 1)!}\int_{u = \log(\eta^2)}^{\log(t^2)}\frac{1}{1 + \lambda^2e^u}\left[\log\left(\frac{t^2}{e^u}\frac{1 +\lambda^2 e^u}{1 + \lambda^2 t^2}\right)\right]^{n - 1}du \\
&= \frac{1}{n!}\int_{\log(\eta^2)}^{\log(t^2)}\frac{d}{du}\left[-\log\left(\frac{t^2}{e^u}\frac{1 + \lambda^2 e^u}{1 + \lambda^2 t^2}\right)\right]^n du ~ = \frac{1}{n!}\left[\log\left(\frac{t^2}{\eta^2}\frac{1 + \lambda^2 \eta^2}{1 + \lambda^2 t^2}\right)\right]^n.
\end{align*}
Leveraging this identity and picking up from the expression for $\phi(\lambda)$ in Equation~\eqref{eq:char_fn}, we have that
\begin{align*}
    \phi(\lambda) &= \frac{\eta^2}{t^2}\sum_{n = 0}^\infty \frac{1}{n!}\left[\log\left(\frac{t^2}{\eta^2}\frac{1 + \lambda^2 \eta^2}{1 + \lambda^2 t^2}\right)\right]^n \\
    &= \frac{\eta^2}{t^2}\exp\left(\log\left(\frac{t^2}{\eta^2}\frac{1 + \lambda^2 \eta^2}{1 + \lambda^2 t^2}\right)\right) ~ = ~ \frac{1 + \lambda^2\eta^2}{1 + \lambda^2 t^2}.
\end{align*}
This proves the desired result.
\end{proof}

The above proof can also be leveraged to show that, for any finite fixed sequence of times $(t_n)_{n \in [K]}$, $(Z(t_1), \dots, Z(t_K))$ has the same distribution as $(\zeta_1, \dots, \zeta_K)$, where $(\zeta_n)_{n \in [K]}$ is the Laplace process associated with times $(t_n)_{n \in [K]}$ as outlined in Equation~\eqref{eq:orig_lap_proc}. This justifies that the process $(Z(t))_{t \geq \eta}$ is in fact a continuous time generalization of the aforementioned discrete time process.

%% file: appendices/bdry.tex
\section{Additional Experimental Details}
\label{app:bdry}

\textbf{Parameter settings:} We set the regularization parameter to be $\lambda = 0.05$ and note that the $\ell_2$ and $\ell_1$-sensitivity for the output perturbation of logistic regression are respectively $\frac{2}{n \lambda}$ and $\frac{2\sqrt{d}}{n \lambda}$. Likewise, for covariance perturbation in ridge regression, the $\ell_2$-sensitivities for privately releasing $X^TX$ and $X^Ty$ are both $2.0$, and the corresponding $\ell_1$-sensitivities for releasing these quantities are $2.0d$ and $2.0\sqrt{d}$ respectively~\citep{ligett2017accuracy, chaudhuri2011differentially}. We set the failure probability for $\BM$ to be $\delta = 10^{-6}$, and in each task map privacy parameters $(\epsilon_n)$ to times $(t_n)$ using the linear privacy boundary $\psi_{a, b}^L$ optimized for tightness at $\epsilon = 0.3$. 

\textbf{Optimizing privacy boundaries:} We provide a high level description of how one may set the parameters associated with the privacy boundaries discussed in Theorem~\ref{thm:priv_bound}. Let us consider the case of the mixture boundary $\psi^{M}_\rho$ for illustrative purposes.

Suppose a data analyst desires that the final level of privacy loss obtained by interacting with the Brownian mechanism should be approximately $\epsilon$. Then, intuitively, the analyst should want to add the variance of the Gaussian noise added to be as small as possible when the privacy boundary takes value $\epsilon$. In mathematical notation, the analyst wants to find a parameter $\rho^\ast$ satisfying
\[
\rho^\ast = \arg \min_\rho (\psi_\rho^{M})^{-1}(\epsilon),
\]
where we note that the inverse function $(\psi_\rho^{M})^{-1}$ exists as $\psi^M_\rho$ is strictly increasing. While this inverse has no closed form in general, the parameter $\rho^\ast$ can be efficiently computed using a few lines of code. A similar, even more straightforward computation can be conducted for the linear privacy boundary.

\textbf{Simulating Noise Reduction Mechanisms:}
We briefly describe how a data analyst can produce samples from the Brownian mechanism and the Laplace noise reduction mechanism. First, since $T_1(x)$ is a constant, we have $\BM_1(x) \sim \calN(f(x), T_1(x))$. Then, given $\BM_{1:m - 1}(x)$, we have $\BM_m(x) \sim \calN\left(f(x) + \frac{T_m(x)}{T_{m - 1}(x)}(B_{T_{m - 1}}(x) - f(x)), \frac{(T_{m - 1}(x) - T_m(x))T_m(x)}{T_{m - 1}(x)}\right)$. Since simulating the Brownian mechnaism only requires normal samples, it can be efficiently computed.

Second, to sample from $\LNR$, one can first generate the the points of arrival of the inhomogeneous Poisson process $(P_t)_{t \geq \eta}$ up to time $T_1(x)$. Let $\calT_1, \dots, \calT_N$ denote these arrival times, where we note that $N$, the number of arrivals up to time $T_1(x)$, is a random variable. Then, one can generate $Y_m \sim \Lap(\calT_m)$ for $m \leq N$. From this information, the process $(Z_t)_{\eta \leq t \leq T_1(x)}$ can be readily computed, as in Equation~\eqref{eq:lap_proc}.